\pdfoutput=1
\documentclass{article}
\usepackage{iclr2027_conference,times}

\iclrfinalcopy

\usepackage[utf8]{inputenc}
\usepackage[T1]{fontenc}
\usepackage{hyperref}
\usepackage{url}
\usepackage{booktabs}
\usepackage{amsfonts}
\usepackage{nicefrac}
\usepackage{microtype}
\usepackage{xcolor}
\usepackage{etoc}
\definecolor{tocblue}{HTML}{1A2F8A}
\colorlet{darkgreen}{green!45!black}
\hypersetup{
  colorlinks,
  citecolor=darkgreen}

\usepackage{graphicx}
\usepackage{wrapfig}
\usepackage{colortbl}
\definecolor{ablneg}{HTML}{D6604D}
\definecolor{ablpos}{HTML}{4393C3}
\newcommand{\abl}[1]{%
  \ifdim #1pt<0pt \cellcolor{ablneg!\fpeval{round(70*sqrt(min(abs(#1),20)/20),1)}}%
  \else \cellcolor{ablpos!\fpeval{round(70*sqrt(min(abs(#1),20)/20),1)}}\fi #1}
\usepackage{enumitem}
\usepackage{multirow}
\usepackage{placeins}

\title{TILE: Bridging LLMs and Table Foundation Models for Tabular Learning with Strings}

\author{%
  Félix Lefebvre\textsuperscript{1} \quad
  David Holzmüller\textsuperscript{1} \quad
  Gioia Blayer\textsuperscript{1} \quad 
  Myung Jun Kim\textsuperscript{2,*} \\
  \bfseries Marine Le Morvan\textsuperscript{1} \quad
  Gaël Varoquaux\textsuperscript{1,3} \\
  \textsuperscript{1}SODA Team, Inria Saclay \quad
  \textsuperscript{2}SAP SE \quad
  \textsuperscript{3}Probabl \quad
  \textsuperscript{*}Work done while at Inria SODA
}

\begin{document}

\etocdepthtag.toc{mtchapter}

\maketitle
\lhead{Preprint.}

\begin{abstract}
Real-world tables are often heterogeneous, combining numerical features with free-form string columns. Table foundation models (TFMs) excel on numerical tables but require strings to be encoded numerically, while large language models (LLMs) hold semantic knowledge but are not designed for tabular predictions. Most existing bridges use LLMs as plug-in encoders for cell strings, yet how to extract representations effectively for tabular learning remains underexplored.
We show that the \emph{interface} between LLMs and tables matters as much as the LLM itself. Applied cell-wise, larger LLMs do not give better features. With our interface, the same LLMs improve steadily with size. TILE (Table-Interfaced LLM Embeddings) combines four ingredients: \emph{row serialization} to provide cross-column context, \emph{intermediate-layer extraction} to avoid layers specialized for next-token prediction, \emph{columnar pooling} to remove mean pooling's length bias, and \emph{by-pass features} to route numbers and categories past the LLM. Paired with TFMs, TILE sets a new state of the art for tables with strings. Across 100 datasets, it beats the strongest baseline on $90\%$ of splits, with gains across 7 LLMs and 10 tabular learners, and ranks first on STRABLE's 108 datasets. Code: \href{https://github.com/soda-inria/tile}{github.com/soda-inria/tile}.
\end{abstract}

\begin{figure}[b!]
    \includegraphics[width=\linewidth]{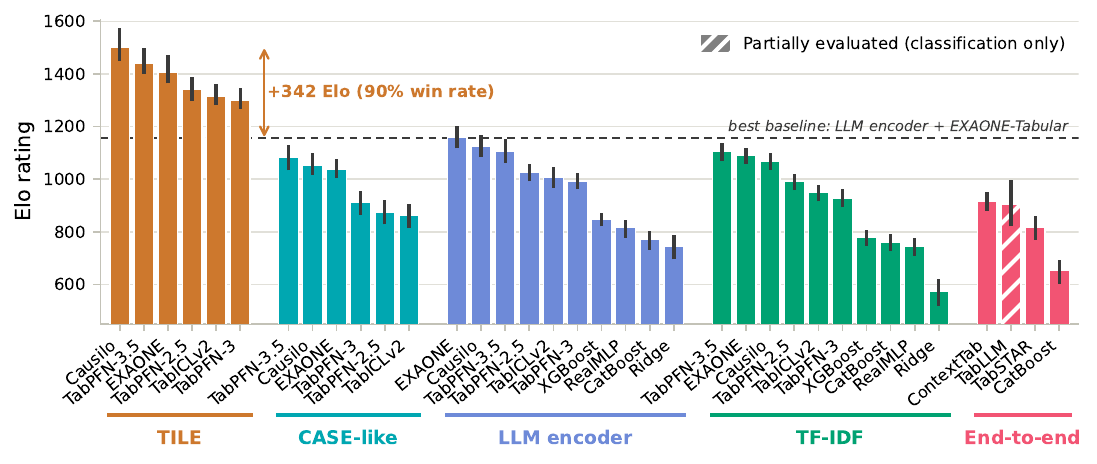}
    \caption{\textbf{TILE outperforms 30 baselines, with every table foundation model.} Elo ratings (\autoref{app:elo}) over 100 regression and classification tasks, at 1,024 train size. TILE uses a Llama-3.1-8B backbone. \emph{LLM encoder} applies the same LLM to each cell in isolation, then a PCA. TabLLM, an LLM fine-tuned to generate tabular predictions, is rated on the 34 datasets where it ran. \emph{CASE-like} is CASE with Gemma-3-4B replacing its unreleased table-tuned backbone (\autoref{app:baselines}).}
    \label{fig:elo}
\end{figure}

\section{Introduction: strings, a new frontier for tabular learning}

From businesses to hospitals, tables underpin critical decisions, making tabular learning---row-wise classification or regression---a cornerstone of applied machine learning \citep{clements2020sequential,pham2026retrieval,van2024tabular}. Yet, modeling real-world tables is challenging because of the heterogeneity of their features. Numerical columns representing quantities sit beside string columns carrying semantics and world knowledge, two modalities historically handled by different learning algorithms \citep{grinsztajn2022tree, shwartz2022tabular}.

Table foundation models, particularly Prior-data Fitted Networks (PFNs) \citep{hollmann2023tabpfn, hollmann2025accurate, qu2025tabicl, qu2026tabiclv2}, now displace the industry-standard tree-based algorithms \citep{chen2016xgboost, prokhorenkova2018catboost} on numerical tables \citep{erickson2025tabarena}. But they are built to model numbers. Standard pipelines therefore prepend a string encoder, such as TF-IDF, that reduces text to surface statistics, discarding both the world knowledge carried by the strings, and the relational context---column headers and neighbouring entries that give cell values their full meaning.

Conversely, LLMs internalize extensive knowledge about the entities mentioned in string cells \citep{petroni2019language, roberts2020much, bordt2024elephants}, but bridging them with tables is not straightforward. Naively applying LLMs to tables forces them to do numerical reasoning, at which they struggle \citep{mirzadeh2025gsmsymbolic, baeumel2025lookahead}, and requires flattening the structure that tabular algorithms exploit \citep{hegselmann2023tabllm, gardner2024large}. Such end-to-end LLMs are in fact outperformed by character $n$-grams (\autoref{fig:elo}), showing that LLMs alone are not enough.

Existing bridges between LLMs and tables force a choice between context and structure. Row- and table-serialization methods \citep{hegselmann2023tabllm, gardner2024large, schindler2025tabgemma} provide context but flatten the table and typically return verbalized answers rather than representations for tabular learners. Cell-wise methods \citep{spinaci2025contexttab, arazi2025tabstar, kim2024carte,kim2025table} embed each string in isolation, which preserves the column structure but gives the LLM short, uncontextualized inputs. The cost of cell isolation becomes clear at scale. Applied cell by cell, larger LLMs do not give better features: across Llama3 models from 1B to 8B parameters the cell-wise encoder barely improves, and across Qwen3 the 0.6B model beats the 8B one (\autoref{fig:serialization}). A lone cell such as ``E.T.'' remains ambiguous, so the knowledge that grows with model size stays locked in. We show that better interfaced to tabular learning, the same LLMs improve steadily with size.

\paragraph{Contributions}
We call the \emph{interface} between an LLM and a table the way the LLM fits into tabular learning: what the LLM is shown, which layer is read, how token embeddings are pooled, and which features are passed to the TFM. Prior work leaves it at its default: isolated cells in, mean-pooled last-layer vectors out. We argue that the interface matters as much as the LLM itself, and turn it into a design space. TILE makes four key choices, each addressing a limitation of current practices.
\textbf{Row serialization} gives every cell its row-level context, addressing the \emph{cell isolation} that leaves cell-wise encoders with strings too short to disambiguate (\autoref{sec:row_serialization}).
\textbf{Intermediate-layer extraction} reads the LLM before the layers specialized for next-token prediction, which do not carry the most transferable semantics (\autoref{sec:probing}).
\textbf{Columnar pooling} restores the column structure that serialization flattens, and removes the \emph{length bias} by which global mean pooling lets verbose columns drown out short informative ones (\autoref{sec:midlayer_pooling}).
Finally, \textbf{by-pass features} route numbers and exact cell identity around the LLM, correcting its numerical imprecision and the \emph{representation drift} that gives one string different embeddings across rows (\autoref{sec:multimodal}).
Together, these choices turn LLM scale into predictive gains: TILE's advantage over the same LLM applied cell-wise grows from $5.3$ points with Qwen3-0.6B to $17.8$ with Qwen3-8B (\autoref{sec:serialization}).
Across 100 datasets, TILE beats 30 baselines, including prior LLM-based methods and the newest TFMs (\autoref{fig:elo}), and ranks first on the external STRABLE benchmark \citep{blayer2026strable}, where TILE with Causilo \citep{cho2026causilo} or TabPFN-3.5 even beats an oracle that selects the best of 63 baselines on each dataset.

\section{Prior work: existing bridges trade context against structure}\label{sec:related}

\paragraph{From tree-based models to foundation models}
Long dominated by gradient-boosted decision trees (GBDTs) \citep{chen2016xgboost, prokhorenkova2018catboost} and their inductive biases for heterogeneous features \citep{grinsztajn2022tree}, tabular learning has seen a resurgence of deep approaches \citep{ye2025revisiting, holzmuller2024better, gorishniy2025tabm} and, with them, table foundation models \citep{hollmann2025accurate, ma2025tabdpt, qu2025tabicl, grinsztajn2025tabpfn, qu2026tabiclv2, grinsztajn2026tabpfn, jager2026tabpfn, eo2026exaone, cho2026causilo, zhang2026limix, tao2026mitra} that beat GBDTs out-of-the-box \citep{erickson2025tabarena}. Rather than fitting one model per dataset, PFNs are pretrained on synthetic or real data and learn \emph{in-context}, processing labelled training rows and unlabelled queries in one forward pass.

\paragraph{Traditional string handling in tabular pipelines}
Categorical variables are usually one-hot or ordinal encoded, and unconstrained strings vectorized by TF-IDF \citep{deerwester1990indexing} or character $n$-grams \citep{cerda2020encoding}, which remains purely surface-level. Studies explore the open problem of language model encodings for text in tables \citep{grinsztajn2023vectorizing, mraz2025towards, lefebvre2025knowledge, koloski2025llm, kolomenko2026embedding, qiang2026tabembed}.

\paragraph{Bridging language models and tabular learners}
Recent work passes either isolated cells or serialized rows to a language model. Cell-level approaches such as ContextTab \citep{spinaci2025contexttab}, TabSTAR \citep{arazi2025tabstar}, CARTE \citep{kim2024carte} and TARTE \citep{kim2025table} rely on relatively small language models \citep{wang2020minilm, wang2022text, bojanowski2017enriching}, which lack the world knowledge of much larger models. Yet cell-wise content is short and uncontextualized, and we show that simply scaling the language model does not help it (\autoref{sec:serialization}).
Row- and table-serialization methods like LIFT \citep{dinh2022lift}, TabLLM \citep{hegselmann2023tabllm}, TabuLa \citep{gardner2024large}, and others \citep{wen2024supervised, schindler2025tabgemma} employ larger models but pay for them with per-dataset fine-tuning or the LLM context limit, and verbalize predictions rather than use a tabular learner. That cost buys little (\autoref{fig:elo}). TaRL \citep{kang2026language} and Latte \citep{shi2025latte} use LLM representations for few-shot row classification, or latent knowledge transfer during training. The recent CASE \citep{schindler2026enhancing} serializes full tables, primes the KV cache of a Tabular Language Model (a fine-tuned Gemma~3) on a sample of training rows, and appends its PCA-reduced last-layer embeddings to the original features. TILE similarly combines LLM representations with a tabular learner, but serializes rows, uses an off-the-shelf LLM, extracts intermediate-layer representations, and pools them column-wise.

\section{TILE: integrating LLMs' representations with TFMs}\label{sec:method}

\begin{figure}[b!]
    \includegraphics[width=\linewidth]{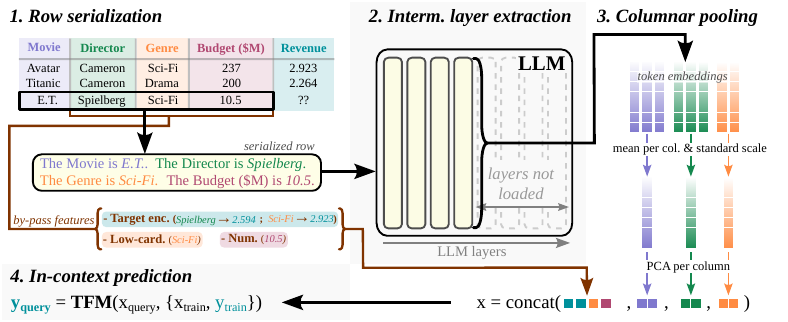}
    \caption{\textbf{The TILE architecture.} Each row is serialized into a sentence and fed to an LLM. Intermediate-layer token embeddings are pooled within string column spans; then standardized and reduced by one PCA per column, both fitted on training rows. The resulting contextualized representation is concatenated with low-cardinality categorical, numerical, and target-encoded categorical features. A TFM makes the final prediction in a single forward pass over the training and query rows.}%
    \label{fig:tile_schema}%
\end{figure}

TILE lets the LLM understand and contextualize the table, and the TFM learn from it (\autoref{fig:tile_schema}; step by step in \autoref{app:pipeline}). Its ingredients follow from two ways in which a table is not a text document.

\paragraph{Columns are heterogeneous, so one path cannot serve them all}
The split between numbers and strings is only the coarsest part of the heterogeneity, and string columns themselves come in kinds. \citet{blayer2026strable} report that across 108 real-world tables about half of the string columns are plain categoricals, a quarter are entity names, a sixth are structured codes, and under a tenth is free text. These make opposite demands. A name such as ``Spielberg'' needs world knowledge, which only a language model supplies; a plain categorical needs exact identity and its association with the target, which an embedding blurs but a target encoding captures exactly; a number needs its value, which an LLM carries imprecisely. A single path for all columns would help some and hurt others; \emph{by-pass features} (\autoref{sec:multimodal}) and a cardinality threshold (\autoref{app:cardinality}) route each column accordingly.

\paragraph{Structure repeats columnwise, so the representation must too}
A text sequence stands alone, whereas a table repeats the same structure across every row: column $j$ carries the same meaning in row 1 and row 1000, and tabular learners exploit exactly that regularity \citep{grinsztajn2022tree}. Serializing a row into a sentence discards this alignment, and pooling that sentence into a single vector discards it again. \emph{Columnar pooling} (\autoref{sec:midlayer_pooling}) restores it.

\subsection{The LLM--table interface: context, depth, and pooling}\label{sec:llm_route}\label{sec:row_serialization}\label{sec:midlayer_pooling}

\paragraph{Row serialization cures cell isolation}
Strings in tables are often very short, such as the few words of a movie title, and short strings are ambiguous: an LLM can resolve them only given context. ``E.T.'' alone could be a time zone; next to ``Spielberg'' in a table about movies, it is the 1982 film (\autoref{fig:tile_schema}). TILE therefore contextualizes each string within its row by row serialization \citep{hegselmann2023tabllm}: each row becomes a sequence of blocks ``The <column name> is <cell value>,'' so that the LLM's own attention conditions each cell on the column headers and on the rest of the row.

\paragraph{Intermediate-layer extraction avoids the output layer's specialization}
Contrary to prior work \citep{hegselmann2023tabllm, gardner2024large}, we do not ask the LLM to verbalize an answer, but extract a representation of its input for a tabular learner. The last layer of a decoder LLM is optimized for next-token prediction, and the first layers still encode surface form; intermediate layers carry more transferable semantics than either end \citep{rogers2020primer, skean2025layer}. TILE reads one layer per LLM, chosen once (14 of 32 for Llama-3.1-8B; \autoref{sec:probing}).

\paragraph{Columnar pooling removes the length bias of mean pooling}
Formally, let $\mathbf{x} = [x_1; \dots; x_M]$ be a table row with $M$ feature columns, serialized into $L$ tokens $\mathbf{t} = (t_1, \dots, t_L)$, and let $H \in \mathbb{R}^{L \times d}$ be the token embeddings $h_i$ at the extraction layer. Standard mean pooling \citep{agarwal2026poolbench} computes $\frac{1}{L} \sum_{i=1}^L h_i$, which is biased toward columns with longer text and dilutes the signal of short but informative ones. Columnar pooling instead gives each string column its own vector. Writing $\mathcal{S}$ for the set of string columns and $\mathcal{I}_j$ for the indices of the tokens of column $j \in \mathcal{S}$ (header and value), the cell embedding is
\begin{equation}\label{eq:columnar_pooling}
    c_j = \frac{1}{|\mathcal{I}_j|} \sum_{i \in \mathcal{I}_j} h_i.
\end{equation}
Numerical columns are serialized, to contextualize the strings, but no embedding is read from their spans, since they reach the learner through the by-pass of \autoref{sec:multimodal}. Each of the $d$ dimensions is then standardized on the training rows, so that none dominates the PCA by scale alone, giving $\tilde{c}_j$.

\paragraph{One PCA per column keeps columns apart and fits the learner}
A cell embedding has far more dimensions than the feature counts TFMs are designed for \citep{tajjar2026pretraining}. TILE fits one PCA $P_j$ per column on the training rows, preserving the alignment that columnar pooling restored:
\begin{equation}\label{eq:row_representation}
    \mathbf{u} = \big[P_j(\tilde{c}_j)\big]_{j \in \mathcal{S}} \in \mathbb{R}^{k \cdot |\mathcal{S}|}.
\end{equation}
TILE uses $k = 30$ components per column by default \citep{grinsztajn2023vectorizing}, rescaled so that the input, with its $p$ by-pass features, fits the window $[a, b]$ the learner handles well---$[200, 400]$ for TabPFN-2.5, $[200, 800]$ for TabICLv2, $[150, 400]$ for TabPFN-3 (\autoref{sec:PCA}): $k = \max\big(1, \min(\lfloor (b-p)/|\mathcal{S}|\rfloor, \max(30, \lceil (a-p)/|\mathcal{S}|\rceil))\big)$. A narrow table thus gets more components per column, and a wide one fewer, which also bounds memory on wide tables.

\subsection{By-pass features address the blind spots of LLM embeddings}\label{sec:multimodal}

The projected representation $\mathbf{u}$ summarizes the row as the LLM reads it, with two blind spots: numbers are carried imprecisely, and \emph{representation drift} makes a string's embedding depend on its row. These columns stay in the sentence, which is what contextualizes the strings, but TILE also routes them around the LLM as \emph{by-pass features}, giving $\tilde{\mathbf{x}} = [\mathbf{u}; \mathbf{x}_{\mathrm{num}}; \mathbf{x}_{\mathrm{cat}}; \mathbf{x}_{\mathrm{te}}]$.

\paragraph{The numerical by-pass preserves exact access to numbers} Table foundation models excel at processing numerical features \citep{erickson2025tabarena}, whereas language models handle numbers unreliably \citep{mirzadeh2025gsmsymbolic, baeumel2025lookahead}. Routing numerical columns only through the LLM would give the TFM a degraded version of the features it handles best. TILE instead also passes them directly to the TFM: the LLM contributes its understanding of the strings, informed by the numbers, while the TFM keeps exact access to the numbers themselves.

\paragraph{The categorical and target-encoding by-passes expose cell identity} Row-level context makes TILE's representations informative, but it also causes representation drift: the same string receives different embeddings in different rows, which hides exact-value repetitions and their association with training targets. TILE therefore gives the TFM two more views of the string columns. String columns with sufficient repetition are target-encoded out of fold (\autoref{app:target_encoding}), and low-cardinality ones (below 50 distinct values; \autoref{app:cardinality}) also enter as native categorical features.

\section{Experimental study: TILE against 30 baselines on 100 tables}\label{sec:experiments}

\subsection{Evaluation protocol: 100 tables with strings}\label{sec:normalization}

\paragraph{Experimental setup}
We evaluate TILE on 100 tables containing strings, drawn from TextTabBench (17), CARTE (51) and WikiDBs (32), with 65 regression and 35 classification tasks (\autoref{app:datasets}). For each dataset, we sample $n_\mathrm{train}$ training and $1{,}024$ test samples (or all remaining ones), over 5 random seeds, and report standard errors of the mean, or 95\% bootstrap intervals for Elo (\autoref{app:elo}).

\paragraph{Metrics and score normalization}
We use $R^2$ (clipped at 0) for regression and ROC-AUC for classification. To aggregate across datasets of varying difficulty, we min-max normalize the scores per dataset and seed \citep{grinsztajn2022tree}, over a fixed pool of 17 baselines (\autoref{app:baselines}) and the three default TILE variants, so that scores remain comparable across figures. Elo ratings (\autoref{fig:elo}) and raw scores (\autoref{app:raw_results}) do not depend on this pool. TILE's few settings---the extraction layer, the feature-capacity window and the cardinality threshold---were set once on this benchmark and kept fixed; each sits on a broad plateau (\autoref{fig:layer-choice}, \autoref{fig:pca-8b}, \autoref{fig:cardinality}), and STRABLE (\autoref{sec:strable}) validates them on separately assembled tables.

\paragraph{Baselines}
We compare TILE with table foundation models---TabPFN-2.5 \citep{grinsztajn2025tabpfn}, TabPFN-3 \citep{grinsztajn2026tabpfn}, TabPFN-3.5 \citep{jager2026tabpfn}, TabICLv2 \citep{qu2026tabiclv2}, EXAONE-Tabular \citep{eo2026exaone} and Causilo \citep{cho2026causilo}---and with tuned RealMLP \citep{holzmuller2024better}, tuned XGBoost \citep{chen2016xgboost}, CatBoost \citep{prokhorenkova2018catboost} and Ridge, each fed with one of two Skrub string encoders. \emph{TF-IDF} (\texttt{StringEncoder}) represents each column with character $n$-gram TF-IDF features followed by SVD. The \emph{LLM encoder} embeds each cell independently, mean-pools its last-layer representations, and reduces each column to 30 PCA components \citep{grinsztajn2023vectorizing}; for a fair comparison, it uses Llama-3.1-8B, TILE's default backbone. We also compare with methods that handle strings themselves: CatBoost, ContextTab \citep{spinaci2025contexttab}, TabSTAR \citep{arazi2025tabstar}, TabLLM \citep{hegselmann2023tabllm} and CASE \citep{schindler2026enhancing}. As CASE's table-tuned backbone is not publicly released, we run it with Gemma-3-4B instead, and call it \emph{CASE-like}. TabPFN-3.5, EXAONE-Tabular and Causilo are not part of the normalization pool. \autoref{app:baselines} details all implementations.

\subsection{TILE establishes a new state-of-the-art for tables with strings}

\begin{table}[t!]
    \caption{\textbf{TILE leads at every train size, on both tasks.} Normalized scores $\times 100$, aggregated over the datasets large enough for each size; counts in parentheses. Because the set of datasets varies, compare scores within columns only. LLM enc.\ is Llama-3.1-8B applied to each cell independently; \emph{CASE-like} is CASE with Gemma-3-4B (\autoref{app:baselines}). Best result in bold, second best underlined. \autoref{tab:large_datasets_full} adds RealMLP, XGBoost, CatBoost and Ridge. \autoref{tab:raw_results} gives raw scores.}
    \label{tab:large_datasets}
    \centering
    \footnotesize
    \setlength{\tabcolsep}{6pt}
    \begin{tabular}{lcccccc}
        & \multicolumn{3}{c}{\textbf{Regression (normalized $R^2$)}} & \multicolumn{3}{c}{\textbf{Classification (normalized AUC)}} \\
        \cmidrule(lr){2-4} \cmidrule(lr){5-7}
        \textbf{Method} & \textbf{1,024 (65)} & \textbf{10k (38)} & \textbf{50k (7)} & \textbf{1,024 (35)} & \textbf{10k (10)} & \textbf{50k (2)} \\
        \midrule
        TILE (TabPFN-2.5) & 90.42 & 93.29 & 92.57 & 87.03 & 88.18 & 79.14 \\
        TILE (TabICLv2) & 88.50 & 92.23 & 92.61 & 88.25 & 85.66 & 80.68 \\
        TILE (TabPFN-3) & 88.36 & 91.94 & 91.89 & 86.91 & 87.32 & 80.22 \\
        TILE (TabPFN-3.5) & \underline{90.59} & \underline{93.66} & \underline{94.43} & \underline{89.59} & \underline{88.28} & \textbf{85.05} \\
        TILE (Causilo) & \textbf{91.76} & \textbf{94.99} & \textbf{95.88} & \textbf{89.97} & \textbf{88.33} & \underline{84.37} \\
        \addlinespace[2pt]
        CASE-like + TabPFN-2.5 & 61.68 & 64.75 & 49.14 & 67.23 & 77.31 & 34.91 \\
        CASE-like + TabICLv2 & 59.23 & 63.05 & 48.99 & 70.13 & 77.02 & 42.19 \\
        CASE-like + TabPFN-3 & 63.10 & 66.10 & 53.76 & 67.24 & 77.98 & 44.08 \\
        CASE-like + TabPFN-3.5 & 70.35 & 73.88 & 57.55 & 78.05 & 85.19 & 62.99 \\
        CASE-like + Causilo & 71.53 & 76.72 & 68.26 & 76.17 & 81.08 & 58.94 \\
        \addlinespace[2pt]
        ContextTab & 70.52 & 78.56 & 66.74 & 69.61 & 80.79 & 49.11 \\
        TabSTAR & 66.67 & 79.23 & 81.33 & 57.19 & 75.98 & 46.12 \\
        CatBoost & 52.16 & 64.19 & 51.96 & 40.82 & 75.93 & 61.80 \\
        \addlinespace[2pt]
        LLM enc. + TabPFN-2.5 & 72.59 & 82.88 & 85.34 & 79.07 & 83.50 & 51.91 \\
        LLM enc. + TabICLv2 & 68.46 & 83.33 & 85.77 & 80.16 & 83.75 & 66.19 \\
        LLM enc. + TabPFN-3 & 73.77 & 78.97 & 83.30 & 73.76 & 77.19 & 49.03 \\
        LLM enc. + TabPFN-3.5 & 75.99 & 85.72 & 85.93 & 80.24 & 83.79 & 60.10 \\
        LLM enc. + Causilo & 77.44 & 84.77 & 88.84 & 80.57 & 83.61 & 63.47 \\
        \addlinespace[2pt]
        TF-IDF + TabPFN-2.5 & 72.58 & 80.11 & 79.62 & 72.21 & 78.65 & 50.54 \\
        TF-IDF + TabICLv2 & 70.74 & 79.00 & 79.92 & 71.42 & 78.11 & 57.15 \\
        TF-IDF + TabPFN-3 & 71.30 & 78.45 & 81.94 & 67.05 & 75.39 & 49.13 \\
        TF-IDF + TabPFN-3.5 & 76.22 & 85.80 & 86.79 & 76.08 & 82.83 & 69.18 \\
        TF-IDF + Causilo & 74.78 & 83.81 & 84.63 & 73.62 & 80.34 & 62.87 \\
        \bottomrule
    \end{tabular}
\end{table}

\paragraph{TILE is the best performer across all training sizes}
\autoref{tab:large_datasets} reports results at three training-set sizes, and \autoref{fig:elo} summarizes the comparison at $n_\mathrm{train}=1{,}024$ with Elo ratings. Every TILE variant ranks above every baseline in all six columns, ahead of tuned gradient-boosted trees, CASE-like, and every cell-wise pipeline built on the same learners. At $1{,}024$ training samples, the best baseline, LLM encoder + Causilo, scores $77.4$ in regression and $80.6$ in classification, against $91.8$ and $90.0$ for TILE (Causilo). In Elo ratings, TILE with Causilo leads the strongest baseline by $342$ points, and beats it on $90\%$ of splits (\autoref{app:elo}). Raw differences are smaller in classification ($89.8$ against $88.6$ AUC), where ROC-AUC saturates near 1 on many tasks (\autoref{app:raw_results}). The regression gap shrinks with training-set size, from $14.3$ points at $1{,}024$ samples to $9.2$ at $10{,}000$ and $7.0$ at $50{,}000$ (over fewer datasets), likely because with more data learners rely less on external knowledge.

\paragraph{Without TILE, LLMs barely improve on TF-IDF} Naive string encodings like TF-IDF, which ignore semantics, are a surprisingly strong baseline: character $n$-grams roughly match the large Llama-3.1-8B applied cell-wise, and beat TabLLM, which fine-tunes an LLM to predict by generating text, for $2.0$ GB of memory against $17.1$ (\autoref{app:memory}). A TFM on plain TF-IDF likewise beats end-to-end ContextTab and TabSTAR. An LLM alone is not enough: its interface matters.

\subsection{External validation: TILE ranks first on STRABLE}\label{sec:strable}

\begin{figure}[t]
    \centering
    \includegraphics[width=0.98\linewidth]{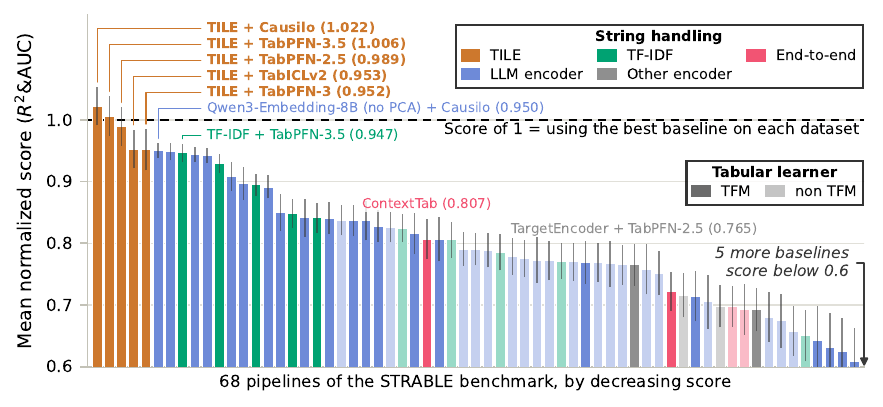}
    \caption{\textbf{TILE ranks first on STRABLE.} Mean normalized score over STRABLE's 108 datasets, under its evaluation protocol. Scores are normalized per dataset over 63 STRABLE baselines, TILE excluded, so a score of 1 amounts to picking the best baseline on each dataset. Colors indicate the string handling; transparent bars mark learners that are not table foundation models, trained from scratch. The five TILE variants rank first, and with Causilo or TabPFN-3.5, TILE scores above 1.}
    \label{fig:strable}
\end{figure}

For a further evaluation varying datasets and protocol, we use STRABLE \citep{blayer2026strable}, a separately assembled benchmark of 108 string-carrying tables. Only 15 of its tables share a source dataset with our benchmark, and those are preprocessed differently (\autoref{app:strable_overlap}).
The five TILE variants take the first five places (\autoref{fig:strable}; raw scores in \autoref{tab:strable_raw}), ahead of the best cell-wise LLM and TF-IDF pipelines. TILE with Causilo or TabPFN-3.5 beats even the per-dataset best baseline, more so with Llama-3.1-70B (\autoref{fig:strable_70b}). The ranking reproduces what we observe on our own datasets: with a TFM, TF-IDF and cell-wise LLMs perform comparably, and end-to-end models trail both. TILE also leads string representations absent from our benchmark (E5, Jasper, All-MiniLM, FastText), providing largely independent evidence that its gains generalize.

\section{Ablations: the interface decides what an LLM contributes}

\subsection{Row context unlocks the benefits of larger LLMs}\label{sec:serialization}

\begin{figure}[t]
    \centering
    \includegraphics[width=0.98\linewidth]{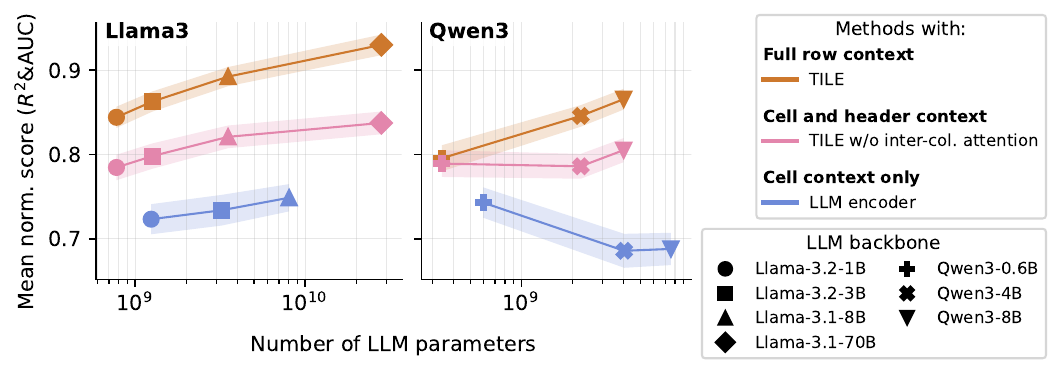}
    \caption{\textbf{Read cell by cell, larger LLMs do not help; with row-level context, they do.} TILE (brown), an ablation masking inter-column attention (pink), and the cell-wise LLM encoder (blue), across four Llama-3 and three Qwen3 backbones. The LLM encoder processes isolated cells, mean-pools last-layer representations, and applies per-column PCA. TILE extracts intermediate representations and thus loads fewer parameters. \autoref{tab:raw_serialization_results} gives raw scores.}
    \label{fig:serialization}
\end{figure}

\paragraph{Scale pays off only through the right interface} \autoref{fig:serialization} compares TILE with the cell-wise LLM encoder across four Llama \citep{grattafiori2024llama} and three Qwen3 models \citep{yang2025qwen3}, spanning two orders of magnitude in size. The cell-wise encoder gains little with scale across Llama, and even declines across Qwen3, where the 0.6B model beats the 8B one. TILE improves steadily with size in both families, so its advantage grows from $5.3$ points for Qwen3-0.6B to $17.8$ for Qwen3-8B. Both pipelines use the same LLMs, pretrained on the same data, and thus the gap reflects only the quality of the interface.
Masking inter-column attention (pink) isolates the role of context, leaving each cell its header but blind to the rest of the row. That ablation still gains from scale, but less: larger models need the full row-level context to reach their potential.

\paragraph{TILE is versatile across LLMs} TILE leads on all seven backbones. For Llama-3.1-70B, the cell-wise encoder does not fit in GPU memory, while truncated TILE does. TILE with Llama-3.1-8B loads about as many parameters as the full Llama-3.2-3B, yet scores $89.2$ against $73.4$ for the latter applied cell-wise (\autoref{app:pareto_tasks}, \autoref{app:memory} give runtime and memory).

\subsection{Intermediate layers and columnar pooling make better representations}\label{sec:probing}

\begin{figure}[t]
    \begin{minipage}[t]{.44\linewidth}\vspace{0pt}
        \includegraphics[width=\linewidth,height=4.4cm,keepaspectratio]{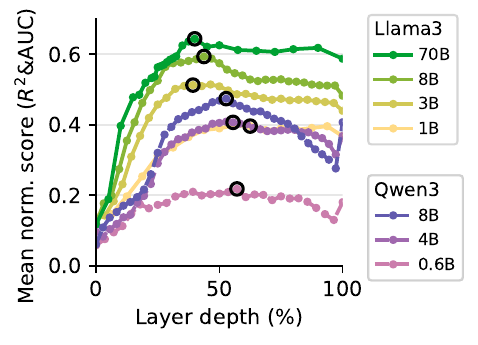}
        \caption{\textbf{Intermediate layers give the best representations, for all 7 backbones.} Columnar-pooled embeddings scored by Ridge (no PCA or by-pass features) across LLM depth, on 27 datasets (\autoref{sec:probing_datasets}). Circles mark TILE's extraction layer.}
        \label{fig:layer-choice}
    \end{minipage}%
    \hfill%
    \begin{minipage}[t]{.53\linewidth}\vspace{0pt}
        \centering
        \includegraphics[width=\linewidth,height=4.4cm,keepaspectratio]{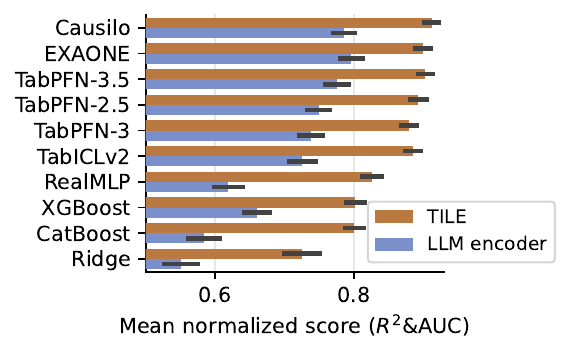}
        \caption{\textbf{TILE boosts every tabular learner.} With TILE, TFMs use their feature-capacity window (\autoref{sec:PCA}); RealMLP, XGBoost and CatBoost use 30 components per column, as the LLM encoder does; Ridge takes the standard-scaled cell embeddings without PCA.}
        \label{fig:learner_ablat}
    \end{minipage}
\end{figure}

We select the extraction layer of each LLM once, on 27 probing datasets. To do so, we extract columnar-pooled representations at every layer, score them with Ridge, without PCA or by-pass features, as a proxy for representation quality, and keep the best layer fixed across all datasets. A single forward pass gives all layers, so this sweep is inexpensive. For all 7 backbones, representation quality climbs steeply through the early layers, plateaus around the middle, and degrades towards the output (\autoref{fig:layer-choice}). Every selected layer lies in the middle third of its network, consistent with \citet{zhang2026attention}, and none is the last; for Llama-3.1-8B, it is layer 14 of 32. On the same probe, columnar pooling beats mean and last-token pooling at every layer (\autoref{app:pooling}).

\subsection{TILE generalizes across tabular learners}

Replacing the table foundation models with alternative learners like RealMLP or Ridge lowers absolute performance (\autoref{fig:learner_ablat}), but with every learner, TILE's representations beat the LLM encoder's, by $10$ to $22$ points. The most recent TFMs, Causilo, TabPFN-3.5 and EXAONE-Tabular, score highest with TILE, showing that its gains add to those of each new generation of learners.

\subsection{Row context and the numerical by-pass carry most of the gain}\label{sec:variations}
\begin{table}[t]
    \caption{\textbf{Ablation study.} Normalized score ($\times 100$) of the default TILE architecture, with Llama-3.1-8B and each of the three default learners, and the difference obtained when removing one of its components or adding a variation to it. The bottom rows give the cell-wise LLM encoder, alone, with the target-encoding and categorical by-passes, or with TILE's feature-capacity window. Normalization puts the regression and classification on a comparable scale; \autoref{tab:raw_ablations} gives raw scores. Cell shading grows with the size of the effect (red: drop, blue: gain). The largest drops result from removing the numerical by-pass or inter-column attention. Columnar pooling cannot be removed without changing the shape of the representation; \autoref{app:pooling} compares pooling strategies.}
    \label{tab:ablations}
    \centering
    \small
    \setlength{\tabcolsep}{6pt}
    \renewcommand{\arraystretch}{1.05}
    \begin{tabular}{@{}lrrrrrr@{}}
        & \multicolumn{2}{c}{\textbf{TabPFN-2.5}} & \multicolumn{2}{c}{\textbf{TabICLv2}} & \multicolumn{2}{c}{\textbf{TabPFN-3}} \\
        \cmidrule(lr){2-3} \cmidrule(lr){4-5} \cmidrule(lr){6-7}
        \textbf{Model} & \textbf{clf} & \textbf{reg} & \textbf{clf} & \textbf{reg} & \textbf{clf} & \textbf{reg} \\
        \midrule
        $-$ numerical by-pass & \abl{-11.29} & \abl{-15.14} & \abl{-11.22} & \abl{-14.35} & \abl{-11.97} & \abl{-16.00} \\
        $-$ inter-column attention & \abl{-2.76} & \abl{-9.51} & \abl{-2.56} & \abl{-6.71} & \abl{-7.34} & \abl{-9.27} \\
        $-$ intermediate layer (last layer) & \abl{-0.61} & \abl{-2.87} & \abl{-0.85} & \abl{-2.51} & \abl{-0.57} & \abl{-2.33} \\
        $-$ feature-capacity window & \abl{+0.02} & \abl{-1.22} & \abl{-0.33} & \abl{-0.42} & \abl{-0.04} & \abl{-0.56} \\
        $-$ numbers in the serialization & \abl{+0.03} & \abl{-0.30} & \abl{-0.70} & \abl{-0.18} & \abl{-0.84} & \abl{-0.35} \\
        $-$ target-encoding by-pass & \abl{+0.06} & \abl{-0.30} & \abl{-0.68} & \abl{+0.20} & \abl{-0.62} & \abl{+0.22} \\
        $-$ standard scaling & \abl{-0.07} & \abl{-0.18} & \abl{-0.27} & \abl{+0.02} & \abl{-0.11} & \abl{-0.18} \\
        $-$ categorical by-pass & \abl{+0.03} & \abl{-0.06} & \abl{-0.04} & \abl{+0.01} & \abl{-0.22} & \abl{-0.06} \\
        \rowcolor{gray!20} TILE (default) & \textbf{87.03} & \textbf{90.42} & \textbf{88.25} & \textbf{88.50} & \textbf{86.91} & \textbf{88.36} \\
        $+$ numerical-cell embeddings & \abl{+0.23} & \abl{-0.31} & \abl{+0.54} & \abl{+0.20} & \abl{-0.21} & \abl{-0.02} \\
        $+$ one-hot instead of native cat. & \abl{-0.13} & \abl{-0.01} & \abl{-0.14} & \abl{-0.05} & \abl{-0.05} & \abl{+0.01} \\
        $+$ repeated serialization & \abl{-1.04} & \abl{+0.16} & \abl{-0.89} & \abl{+0.25} & \abl{-1.22} & \abl{+0.05} \\
        $+$ mirrored column order & \abl{-0.22} & \abl{-0.75} & \abl{-0.08} & \abl{-0.25} & \abl{-1.13} & \abl{-0.95} \\
        $+$ target name & \abl{-0.61} & \abl{-0.66} & \abl{-0.74} & \abl{-0.59} & \abl{-0.38} & \abl{-0.71} \\
        $+$ uninformative column names & \abl{-2.11} & \abl{-2.90} & \abl{-2.69} & \abl{-2.90} & \abl{-2.88} & \abl{-2.85} \\
        \midrule
        LLM encoder (\texttt{skrub}) & \abl{-7.96} & \abl{-17.83} & \abl{-8.09} & \abl{-20.04} & \abl{-13.15} & \abl{-14.59} \\
        LLM encoder $+$ by-pass features & \abl{-7.17} & \abl{-15.56} & \abl{-7.87} & \abl{-18.38} & \abl{-14.24} & \abl{-16.23} \\
        LLM encoder $+$ feature-capacity window & \abl{-8.06} & \abl{-16.97} & \abl{-7.96} & \abl{-20.20} & \abl{-16.34} & \abl{-18.36} \\
        \bottomrule
    \end{tabular}
\end{table}

\autoref{tab:ablations} measures the contribution of each component of TILE, and evaluates variations that were not retained. The rows above the default remove a component; those below add to or modify it; the bottom rows give the cell-wise LLM encoder. The three learners agree; we quote TabPFN-2.5.

\paragraph{The semantic and the by-pass paths compound} Removing the numerical by-pass is the most damaging ablation, even though the LLM still reads the numbers. Conversely, omitting numerical values from the serialization also hurts, even though those columns reach the learner through the by-pass. Neither path substitutes for the other: the LLM needs the numbers to interpret the strings, and the learner needs them exactly for its predictions. Pooling embeddings of the numerical cells as well, however, does not help consistently: within the TFM's feature-capacity window, they come at the cost of compressing the string embeddings further.
On the semantic path, masking inter-column attention causes the second-largest drop, and reading the last layer instead of layer 14 costs $0.6$ points in classification and $2.9$ in regression, confirming the probing results of \autoref{sec:probing} on the full pipeline. Standard scaling and the target-encoding by-pass each bring less than a point, and the feature-capacity window $1.2$ points in regression.
None of these components alone explains the gap to the cell-wise LLM encoder, which shares TILE's numerical by-pass: masking inter-column attention, the largest, costs $2.8$ (classification) and $9.5$ (regression) points, against a gap of $8.0$ and $17.8$. Giving the LLM encoder TILE's other by-passes narrows the gap only to $7.2$ and $15.6$, and sizing its PCA with the feature-capacity window leaves it at $8.1$ and $17.0$ (both widen it with TabPFN-3).

\paragraph{TILE uses table-level context}
Replacing informative column headers with ``Col 1'', ``Col 2'', etc., or reversing the column order degrades performance, showing that TILE exploits both the column names and their arrangement. The effect of ordering is likely due to the causal mask of decoder LLMs, under which each cell attends only to preceding columns. As real tables often put identifying columns first, reversing them may deprive entries of informative context.

\paragraph{Richer prompts do not pay off} Prepending the target column name (``What is the value of <target column name> given that:'') is meant to inform the model of the task, but degrades performance. Repeating the serialized row twice and reading the embeddings from the second copy, inspired by prompt repetition in LLMs \citep{springer2025repetition, leviathan2025prompt}, lets every column attend to the whole row despite the causal mask. However, it hurts in classification, barely changes regression, and doubles the input length.

\section{Discussion: unlocking LLM world knowledge for tables}\label{sec:conclusion}

\paragraph{Building the right interface between LLMs and tables} Pairing an LLM with a strong tabular learner, TILE turns strings into predictive signal, with sizable gains over the state of the art (\autoref{fig:elo}, \autoref{tab:large_datasets}). Our results suggest that what has held back LLMs on tables is their interface to the table. With row-level context and a well-chosen layer, TILE makes LLMs' gains grow with their size (\autoref{fig:serialization}). TILE is versatile and improves every LLM backbone (\autoref{fig:serialization}) and tabular learner (\autoref{fig:learner_ablat}) we evaluate, and gains from advances in both, as the newest TFMs show.

\paragraph{Limitations and future work} TILE targets tables whose strings carry predictive signal; on purely numerical tables, a standalone TFM suffices. Its LLM embedding stage needs a GPU, though truncation nearly halves its weights (\autoref{app:memory}). Row serialization is bounded by the LLM's context window; this was never a constraint in our benchmark, whose longest serialized row is $3{,}425$ tokens against the $128$k-token context of Llama-3.1-8B, but very wide tables with long texts would need chunking. Finally, the causal mask of decoder LLMs makes representations depend on column order, which bidirectional encoders \citep{behnamghader2024llm2vec} would avoid: a natural next step.

\subsection*{AI use statement}

In this work, we used generative AI tools as general-purpose assistants for writing and coding. For writing, they were used to polish prose and improve the clarity of text drafted by the authors; all scientific claims, experimental results and their interpretation were written and verified by the authors. For coding, they were used to assist with implementation and with the scripts that produce the figures and tables; all such code was read, tested and validated by the authors against the experimental protocol described in \autoref{sec:experiments} and \autoref{app:baselines}. We did not use generative AI tools for research ideation, for the design of the method, or for the literature review, and they are not part of the TILE pipeline beyond the publicly released language models that it reads representations from (\autoref{app:licenses}). We have reviewed all AI-assisted work and take responsibility for the final content of this work, including its text, claims and artifacts.

\subsection*{Ethics statement}

By relying on LLMs, TILE may internalize and propagate the societal biases present in their pretraining data \citep{bender2021dangers, hu2025generative}. In sensitive applications such as hiring, lending, or criminal justice, these encoded biases could produce unfair predictions or exacerbate existing demographic disparities, making pre-deployment audits crucial. Conversely, by improving predictive performance on heterogeneous tables, TILE could positively impact critical tabular-driven domains, such as healthcare, where patient records frequently intermix clinical measurements with free-text notes or diagnoses \citep{johnson2016mimic, miotto2016deep}.

\subsection*{Reproducibility statement}

TILE is built entirely from publicly released artifacts, and we have aimed to make every step of our evaluation reproducible. The code is available at \url{https://github.com/soda-inria/tile}, and the preprocessed datasets of the benchmark at \url{https://huggingface.co/datasets/inria-soda/TILE-benchmark}. \autoref{sec:method} specifies the architecture, and \autoref{app:licenses} lists every language model, tabular learner and software package we use, together with its source and license. The benchmark is described in \autoref{app:datasets}, with per-dataset statistics and the preprocessing applied to each table; the training-set sizes, splits, metrics and aggregation procedure are given in \autoref{sec:experiments}, and the Elo ratings of \autoref{fig:elo} are defined in \autoref{app:elo}. Every baseline, with its implementation, version and hyperparameter search space, is documented in \autoref{app:baselines}, and \autoref{app:target_encoding} details the target encoding of categorical string columns. Raw, unaggregated scores for all methods are reported in \autoref{app:raw_results}, and the compute and memory budget of each stage in \autoref{app:compute_resources}.

\subsubsection*{Acknowledgments}
This work was granted access to the HPC resources of IDRIS under the allocation 2025-AD011016808 made by GENCI. GV acknowledges support from ANR via grant TaFoMo (ANR-25-CE23-1822). This work is partly supported by Hi! PARIS and ANR/France 2030 program (ANR-23-IACL-0005).

\bibliographystyle{iclr2027_conference}
\bibliography{references}

\appendix

\clearpage
\etocdepthtag.toc{mtappendix}
\counterwithin{figure}{section}
\counterwithin{table}{section}

\begingroup
  \etocsettagdepth{mtchapter}{none}
  \etocsettagdepth{mtappendix}{subsection}
  \setlength{\parskip}{0pt}

  \etocsettocstyle{%
    \noindent{\LARGE Appendices}\par\vspace{1.4em}%
    \noindent{\large Contents}\par\vspace{0.8em}}{}

  \etocsetstyle{section}
    {}
    {\addvspace{1.3ex}}
    {\noindent
     \etoclink{\color{tocblue}\makebox[1.5em][l]{\etocthenumber}\etocthename}%
     \hfill\makebox[2em][r]{\etocthepage}\par}
    {}

  \etocsetstyle{subsection}
    {\addvspace{0.3ex}}
    {}
    {\noindent\hspace*{1.5em}%
     \etoclink{\color{tocblue}\makebox[2.4em][l]{\etocthenumber}\etocthename}%
     \nobreak\leaders\hbox to 0.9em{\hss.\hss}\hfill
     \makebox[2em][r]{\etocthepage}\par\addvspace{0.35ex}}
    {}

  \tableofcontents
\endgroup

\clearpage

\section{Evaluation datasets}\label{app:datasets}

\subsection{Data sources}
To evaluate tabular learning with strings, we assemble a diverse benchmark of 100 tabular datasets from three sources with distinct characteristics and inclusion biases: TextTabBench \citep{mraz2025towards}, CARTE \citep{kim2024carte}, and WikiDBs \citep{vogel2024wikidbs}.

TextTabBench and CARTE are established benchmarks for tabular learning, providing real-world tables with varied text features, from short entity names to longer descriptions. Each table is associated with a predefined prediction task (regression, binary, or multi-class classification). WikiDBs is a large corpus of over 1.6 million semi-synthetic tables generated from Wikidata. To create meaningful tasks from this source, we first filtered for tables with at least 1,200 rows, then manually curated a subset of 32 tables for which we could define a relevant prediction problem. \autoref{tab:task_distribution} summarizes the final distribution of tasks across the three sources. \autoref{tab:ttb_datasets}, \autoref{tab:carte_datasets} and \autoref{tab:wdbs_datasets} provide further details on each individual dataset.

\begin{table}[h]
    \small
    \centering
    \caption{Task distribution across sources.}
    \label{tab:task_distribution}
    \begin{tabular}{lcccc}
    \toprule
    \textbf{Source} & \textbf{b-clf} & \hspace*{-.5ex}\textbf{m-clf}\hspace*{-.5ex} & \textbf{reg} & \textbf{Total}\hspace*{-1ex} \\
    \midrule
    TextTabBench\hspace*{-2ex} & 5 & 2 & 10 & \textbf{17} \\
    CARTE & 11 & - & 40 & \textbf{51} \\
    WikiDBs & 1 & 16 & 15 & \textbf{32} \\
    \midrule
    \textbf{Total} & \textbf{17} & \textbf{18} & \textbf{65} & \textbf{100} \\
    \bottomrule
    \end{tabular}
\end{table}

\begin{table}[b]
\centering
\caption{Overview of TextTabBench datasets used in our benchmark. Table statistics after preprocessing.}
\label{tab:ttb_datasets}
\begin{tabular}{llrrrr}
\toprule
\textbf{Dataset} & \textbf{Task} & \textbf{\# rows} & \multicolumn{2}{c}{\textbf{\# columns}} & \textbf{\# classes} \\
\cmidrule(lr){4-5}
& & & \textit{cat.} & \textit{num.} & \\
\midrule
Diabetes & b-clf & 17,000 & 4 & 12 & 2 \\
Job Frauds & b-clf & 1,732 & 11 & 3 & 2 \\
Kickstarter & b-clf & 18,720 & 9 & 8 & 2 \\
Lending Club & b-clf & 11,254 & 12 & 15 & 2 \\
Osha Accidents & b-clf & 3,598 & 15 & 3 & 2 \\
\midrule
Customer Complaints & m-clf & 1,384 & 8 & 2 & 4 \\
Spotify & m-clf & 10,000 & 3 & 15 & 10 \\
\midrule
\midrule
Airbnb & reg & 3,818 & 32 & 23 & - \\
Beer & reg & 2,914 & 5 & 15 & - \\
California Houses & reg & 11,349 & 13 & 16 & - \\
Covid Trials & reg & 1,165 & 13 & 2 & - \\
Insurance Complaints & reg & 37,484 & 8 & 2 & - \\
IT Salary & reg & 1,253 & 16 & 2 & - \\
Mercari & reg & 12,000 & 4 & 2 & - \\
San Francisco Permits & reg & 183,794 & 12 & 16 & - \\
Stack Overflow & reg & 19,427 & 89 & 13 & - \\
Wine & reg & 1,281 & 12 & 3 & - \\
\bottomrule
\end{tabular}
\end{table}

\begin{table}[t]
\centering
\caption{Overview of CARTE datasets used in our benchmark. Table statistics after preprocessing.}
\label{tab:carte_datasets}
\small
\begin{tabular}{llrrrr}
\toprule
\textbf{Dataset} & \textbf{Task} & \textbf{\# rows} & \multicolumn{2}{c}{\textbf{\# columns}} & \textbf{\# classes} \\
\cmidrule(lr){4-5}
& & & \textit{cat.} & \textit{num.} & \\
\midrule
Chocolate Bar Ratings & b-clf & 2,218 & 6 & 5 & 2 \\
Coffee Ratings & b-clf & 1,670 & 8 & - & 2 \\
Michelin & b-clf & 6,774 & 5 & 2 & 2 \\
NBA Draft & b-clf & 1,550 & 4 & 5 & 2 \\
Ramen Ratings & b-clf & 3,726 & 4 & - & 2 \\
Roger Ebert & b-clf & 2,668 & 5 & 5 & 2 \\
Spotify & b-clf & 41,096 & 7 & 11 & 2 \\
US Accidents Severity & b-clf & 20,930 & 9 & 4 & 2 \\
Whisky & b-clf & 1,788 & 6 & - & 2 \\
Yelp & b-clf & 60,088 & 8 & 4 & 2 \\
Zomato & b-clf & 60,302 & 7 & 1 & 2 \\
\midrule
\midrule
Movies & reg & 7,224 & 7 & 9 & - \\
US Accidents Counts & reg & 22,623 & 6 & - & - \\
US Presidential & reg & 19,857 & 6 & - & - \\
Anime Planet & reg & 14,391 & 6 & 10 & - \\
Babies R Us & reg & 5,085 & 4 & - & - \\
Beer Ratings & reg & 3,197 & 5 & 14 & - \\
Bikedekho & reg & 4,786 & 5 & 5 & - \\
Bikewale & reg & 8,992 & 5 & 5 & - \\
Buy Buy Baby & reg & 10,718 & 4 & - & - \\
Cardekho & reg & 37,813 & 13 & 6 & - \\
Clear Corpus & reg & 4,724 & 10 & 19 & - \\
Company Employees & reg & 10,941 & 7 & - & - \\
Employee Remuneration & reg & 35,396 & 2 & 5 & - \\
Employee Salaries & reg & 9,211 & 6 & 8 & - \\
Fifa22 Players & reg & 18,085 & 9 & 18 & - \\
Filmtv Movies & reg & 41,205 & 6 & 6 & - \\
Journal JCR & reg & 9,615 & 4 & 5 & - \\
Journal SJR & reg & 27,931 & 9 & - & - \\
Japanese Anime & reg & 15,535 & 11 & 5 & - \\
K-Drama & reg & 1,239 & 8 & 4 & - \\
ML/DS Salaries & reg & 10,456 & 7 & 4 & - \\
Museums & reg & 11,467 & 14 & 2 & - \\
Mydramalist & reg & 3,400 & 10 & 3 & - \\
Prescription Drugs & reg & 1,714 & 5 & 5 & - \\
Rotten Tomatoes & reg & 7,158 & 10 & 6 & - \\
Used Cars 24 & reg & 5,918 & 6 & 5 & - \\
Used Cars Benz Italy & reg & 16,391 & 5 & 2 & - \\
UsedCars.com & reg & 4,009 & 8 & 5 & - \\
Used Cars Pakistan & reg & 72,655 & 4 & 6 & - \\
Used Cars Saudi Arabia & reg & 5,507 & 7 & 6 & - \\
Videogame Sales & reg & 16,410 & 4 & 4 & - \\
Wikiliq Beer & reg & 13,461 & 7 & 2 & - \\
Wikiliq Spirit & reg & 12,275 & 5 & 2 & - \\
Wina Poland & reg & 2,247 & 12 & 6 & - \\
Wine.com Prices & reg & 15,254 & 6 & 3 & - \\
Wine.com Ratings & reg & 4,095 & 6 & 3 & - \\
WineEnthusiasts Prices & reg & 120,975 & 8 & 1 & - \\
WineEnthusiasts Ratings & reg & 129,971 & 8 & 1 & - \\
WineVivino Price & reg & 13,834 & 5 & 2 & - \\
WineVivino Rating & reg & 13,834 & 6 & 2 & - \\
\bottomrule
\end{tabular}
\end{table}

\begin{table}[t]
\centering
\caption{Overview of WikiDBs datasets used in our benchmark. Table statistics after preprocessing.}
\label{tab:wdbs_datasets}
\begin{tabular}{llrrrr}
\toprule
\textbf{Dataset} & \textbf{Task} & \textbf{\# rows} & \multicolumn{2}{c}{\textbf{\# columns}} & \textbf{\# classes} \\
\cmidrule(lr){4-5}
& & & \textit{cat.} & \textit{num.} & \\
\midrule
CC Authors & b-clf & 16,224 & 7 & 1 & 2 \\
\midrule
Defenders & m-clf & 18,610 & 9 & - & 10 \\
Philosophers & m-clf & 4,230 & 7 & - & 10 \\
US Music Albums & m-clf & 3,270 & 10 & 1 & 10 \\
Artist Copyrights & m-clf & 2,000 & 9 & 1 & 10 \\
Artworks Catalog & m-clf & 1,210 & 7 & 2 & 10 \\
Forward Players & m-clf & 1,400 & 10 & - & 10 \\
Geographers & m-clf & 1,130 & 8 & - & 10 \\
Island & m-clf & 19,650 & 2 & 2 & 10 \\
Magic Narratives & m-clf & 1,062 & 4 & - & 9 \\
Museums & m-clf & 9,550 & 4 & 2 & 10 \\
Noble Individuals & m-clf & 1,400 & 9 & - & 10 \\
Notable Trees & m-clf & 1,408 & 4 & 2 & 8 \\
Sculptures & m-clf & 3,720 & 6 & - & 10 \\
Spring Locations & m-clf & 5,930 & 1 & 2 & 10 \\
State Schools & m-clf & 2,800 & 2 & 2 & 10 \\
Transport Stations & m-clf & 4,640 & 7 & 2 & 10 \\
\midrule
\midrule
Business Locations & reg & 16,821 & 4 & 4 & - \\
Dissolved Municipalities & reg & 13,462 & 6 & 2 & - \\
Geopolitical Regions & reg & 1,114 & 6 & 3 & - \\
Historical Figures & reg & 11,260 & 11 & - & - \\
Municipal District Capitals & reg & 1,658 & 5 & 3 & - \\
Poets & reg & 60,240 & 10 & - & - \\
Territorial Entities & reg & 36,717 & 7 & 4 & - \\
WWI Personnel & reg & 30,675 & 11 & - & - \\
Artworks Inventory & reg & 10,635 & 5 & 1 & - \\
Drawings Catalog & reg & 63,130 & 8 & 1 & - \\
Eclipsing Binary Stars & reg & 297,934 & 6 & 2 & - \\
Registered Ships & reg & 4,644 & 6 & 3 & - \\
Research Articles & reg & 6,962 & 6 & 2 & - \\
Research Article Citations & reg & 4,115 & 9 & - & - \\
Ukrainian Villages & reg & 21,355 & 3 & 3 & - \\
\bottomrule
\end{tabular}
\end{table}

\subsection{Data preprocessing}

We adopt the original preprocessing from TextTabBench and CARTE. For WikiDBs, we apply a procedure similar to TextTabBench. We also ensure that multi-class classification tasks have at most 10 classes, each with at least 105 samples. For all 100 datasets, we then apply the following preprocessing pipeline: (1) we log-transform regression targets with wide-ranging distributions; (2) we downsample majority classes in multi-class problems to create balanced datasets; and (3) we discard any table with fewer than 1,050 rows post-processing to ensure sufficient data for evaluation. We also exclude one dataset from TextTabBench with excessively long text entries, as detailed in \autoref{sec:excluded_dataset}.

\subsection{Data characteristics}

\autoref{tab:dataset_features} reports aggregated features across datasets, broken down by data source. In particular, it shows that TextTabBench tables tend to contain longer text fields while CARTE and WikiDBs tables gather shorter string entries.

\begin{table}[t!]
    \caption{\textbf{Aggregated features of tabular datasets across sources.} The cardinality is computed over 1,024 rows. ``String similarity" represents cosine similarity of TF-IDF across rows.}
    \label{tab:dataset_features}
    \centering
    \begin{tabular}{lrrr}
    \toprule
    & \textbf{TextTabBench} & \textbf{CARTE} & \textbf{WikiDBs} \\
    \midrule
    \# columns & 15.7 & 6.8 & 6.7 \\
    cardinality & 286.4 & 371.4 & 463.7 \\
    string length & 975.3 & 298.8 & 203.6 \\
    string similarity & 0.16 & 0.10 & 0.08 \\
    \bottomrule
    \end{tabular}
\end{table}

\subsection{Excluded dataset}\label{sec:excluded_dataset}

The dataset excluded from our benchmark is the OkCupid table of TextTabBench, a multi-class classification task with 59,946 rows, 28 string columns and very long text entries. The exclusion is a constraint of the baselines, not of our method: the cell-wise LLM encoder runs out of memory on it, so no LLM-encoder pipeline ran on this dataset, and including it would leave holes in the comparison. TILE processes it without difficulty, as do the TF-IDF pipelines.

For completeness, \autoref{tab:okcupid} reports the methods that do run on all 101 tables. Adding this dataset leaves the ranking unchanged. In absolute terms it is a hard task for every method, with raw ROC AUC scores ranging from 65.0 to 76.0, the best being TILE with TabPFN-3.5 (75.7 with Causilo).

\begin{table}[h]
\begin{minipage}{.36\linewidth}
    \caption{\textbf{Adding the excluded dataset.} Normalized scores over the excluded table alone, over the benchmark, and over their union. Normalization is over the 16 methods that ran on all 101 tables, a smaller pool than in the rest of the paper, so the ranking is what should be read.}
    \label{tab:okcupid}
\end{minipage}%
\hfill%
\begin{minipage}{.6\linewidth}
    \centering
    \small
    \setlength{\tabcolsep}{4pt}
    \begin{tabular}{lccc}
        \toprule
        & \textbf{OkCupid} & \textbf{Benchmark} & \textbf{Both} \\
        \textbf{Method} & \textbf{(1)} & \textbf{(100)} & \textbf{(101)} \\
        \midrule
        TILE (Causilo) & 88.2 & \textbf{87.6} & \textbf{87.6} \\
        TILE (TabPFN-3.5) & \textbf{89.9} & 86.2 & 86.3 \\
        TILE (EXAONE-Tabular) & 59.5 & 85.8 & 85.6 \\
        TILE (TabPFN-2.5) & 55.3 & 84.9 & 84.6 \\
        TILE (TabICLv2) & 84.2 & 83.8 & 83.8 \\
        TILE (TabPFN-3) & 65.4 & 83.1 & 82.9 \\
        TF-IDF + TabPFN-3.5 & 67.8 & 69.2 & 69.2 \\
        TF-IDF + EXAONE-Tabular & 36.0 & 68.4 & 68.0 \\
        TF-IDF + Causilo & 59.0 & 67.4 & 67.3 \\
        TF-IDF + TabPFN-2.5 & 56.4 & 64.4 & 64.3 \\
        TF-IDF + TabICLv2 & 65.2 & 62.9 & 62.9 \\
        ContextTab & 65.6 & 60.7 & 60.7 \\
        TabSTAR & 22.6 & 53.8 & 53.5 \\
        TF-IDF + XGBoost & 40.5 & 53.0 & 52.8 \\
        TF-IDF + RealMLP & 29.4 & 50.9 & 50.7 \\
        TF-IDF + Ridge & 51.2 & 25.4 & 25.6 \\
        \bottomrule
    \end{tabular}
\end{minipage}
\end{table}

\subsection{Probing datasets used to set TILE's hyperparameters}\label{sec:probing_datasets}
All of TILE's hyperparameters are set on a subset of 27 datasets: the extraction layer of each LLM (\autoref{fig:layer-choice}), the feature-capacity window of each learner (\autoref{sec:PCA}) and the cardinality threshold of the categorical by-pass (\autoref{app:cardinality}). The pooling comparison of \autoref{fig:pooling-8b} uses them too. The datasets used are the following:

\begin{itemize}[noitemsep,topsep=2pt,parsep=0pt,partopsep=0pt,leftmargin=3ex]
  \item \textbf{CARTE datasets:} NBA Draft, Prescription Drugs, Coffee Ratings, Ramen Ratings, Chocolate Bar Ratings, Roger Ebert, UsedCars.com, Used Cars 24, Michelin, Beer Ratings, Company Employees, Journal JCR, Videogame Sales, Filmtv Movies, Wine Vivino Price, Zomato, Yelp.

  \item \textbf{TextTabBench datasets:} Beer, Mercari.

  \item \textbf{WikiDBs datasets:} Magic Narrative, Geopolitical Regions, Sculptures, Research Articles, Ukrainian villages, Business Locations, Territorial Entities, Drawings Catalog.
\end{itemize}

\subsection{Overlap with STRABLE}\label{app:strable_overlap}

STRABLE \citep{blayer2026strable} and our benchmark were assembled separately, but both draw on the small pool of public tables with meaningful text, so some sources appear in both. Of STRABLE's 108 tables, 15 (about $14\%$) share a source dataset with CARTE or TextTabBench; the remaining 93 are unique to STRABLE.

\paragraph{Shared sources} Ten of the shared tables come from CARTE: AI/ML salaries, beer ratings, chocolate bar ratings, CLEAR corpus, SCImago journal ranking, Michelin, ramen ratings, whisky, Wine Enthusiast, and Yelp. A sixteenth match is possible if CARTE's Kaggle museum table, a repackaging of the IMLS Museum Universe file, is counted. Six come from TextTabBench: beer ratings (also in CARTE), CFPB complaints, Kickstarter, Mercari, OSHA accidents, and San Francisco building permits. Two of the TextTabBench-shared sources, OSHA and San Francisco permits, are paired with a different prediction target in STRABLE: injury severity and construction cost, rather than task regularity and approval time. Several nominal matches are distinct datasets: STRABLE's California Housing is the 1990 census table, not TextTabBench's text-rich listings, and TextTabBench's wine and Spotify tables have no STRABLE counterpart.

\paragraph{Shared tables are not identical copies} STRABLE applies uniform minimal preprocessing, with no dataset-specific feature engineering. It keeps the most recent snapshot of each source and applies a skewness-minimizing target transform. CARTE, by contrast, applies hand-crafted per-dataset transformations, such as recoding whisky price brackets and binarizing its target.

\paragraph{The two benchmarks are largely, though not fully, independent} On STRABLE's side, the overlap is small relative to its size and concentrated in its Food, Commerce, and Education domains. On ours, the shared sources account for 17 of the 100 tables, or 18 if the museum table is counted: 11 of the 51 CARTE tables, as Wine Enthusiast gives both a price and a rating table, and 6 of the 17 TextTabBench tables. This overlap reflects the scarcity of public tables with meaningful text rather than a derivative design. Results on STRABLE can thus be read as largely complementary evidence to those on our benchmark, with the caveat that a model's gains on the ${\sim}15$ shared sources should not be counted as independent confirmation across benchmarks.

\clearpage
\section{Elo ratings}\label{app:elo}

\autoref{fig:elo} summarizes the benchmark with Elo ratings, computed with the implementation of TabArena~\citep{erickson2025tabarena}, which follows Chatbot Arena~\citep{chiang2024chatbot}. \autoref{tab:elo} lists the rating of every method, with its confidence interval.

\paragraph{Pairwise comparisons}
Each method $m$ is evaluated on dataset $d$ with 5 random train/test splits (seeds) $s$, the same for every method, giving an error $e_{m,d,s} = 1 - R^2$ for regression and $1 - \mathrm{ROC\ AUC}$ for classification. The unit of comparison is a \emph{battle}: for every dataset $d$, seed $s$ and pair of methods $(a,b)$ that were both run on $(d,s)$, method $a$ wins if $e_{a,d,s} < e_{b,d,s}$, loses if $e_{a,d,s} > e_{b,d,s}$, and a tie counts as half a win for each (ties occur in fewer than $0.01\%$ of battles). Each battle has weight $1/5$, so that every dataset has the same total weight. Because a battle only compares two methods on the same split, regression and classification datasets can be pooled into a single rating without ever comparing $R^2$ with ROC AUC. Over the 36 methods and 100 datasets this gives 303{,}345 battles.

\paragraph{Rating model}
The ratings $r_m$ are the maximum-likelihood estimate of a Bradley--Terry model~\citep{bradley1952rank} on the Elo scale,
\begin{equation}
  P(a \mbox{ beats } b) = \frac{1}{1 + 10^{-(r_a - r_b)/400}},
  \label{eq:elo}
\end{equation}
fit by weighted logistic regression over all battles at once (with negligible regularization, $C = 10^6$). Unlike the online Elo update, this estimate uses no K-factor and does not depend on the order of the battles: it is the fixed point that online Elo approaches with a vanishing step size. Ratings are only defined up to a common shift, which we fix so that the mean rating over methods is 1000.

\paragraph{Interpretation}
By Equation~\ref{eq:elo}, only differences between ratings matter. A gap of $\Delta$ points translates into an expected head-to-head win probability: $\Delta = 100$ gives $64\%$, $\Delta = 200$ gives $76\%$ and $\Delta = 400$ gives $91\%$. The $+342$ Elo gap between TILE (Causilo) and the strongest baseline (LLM encoder + EXAONE-Tabular) thus means that TILE is expected to beat that baseline on $87.7\%$ of splits. On the 500 splits where both were run, it actually wins on $89.9\%$ (449 wins and 1 tie, counted as half a win), and $89.9\%$ as well when averaged per dataset. This strongest baseline does not appear in \autoref{tab:large_datasets}: because of compute constraints, we ran LLM encoder + EXAONE-Tabular only at $n_\mathrm{train}=1{,}024$, the training size of the Elo ratings and of the per-dataset results of \autoref{app:raw_results}, and not yet at $10{,}000$ and $50{,}000$.

\paragraph{Confidence intervals}
Error bars are 95\% percentile bootstrap intervals. We resample the 100 datasets with replacement (all seeds of a drawn dataset are kept together), refit the ratings on the battles of the resample, and repeat this 200 times (bootstrap random seed 0). The bars go from the 2.5\% to the 97.5\% quantile of the bootstrap ratings and are drawn around the point estimate fitted on the full benchmark, which is why they can be asymmetric. Resampling datasets rather than individual battles accounts for the correlation between battles on the same dataset, so the intervals reflect uncertainty over the choice of datasets. Per-method intervals do not directly give the uncertainty of a difference, so we bootstrap the gap itself with the same resamples: the gap between TILE (Causilo) and LLM encoder + EXAONE-Tabular has a 95\% interval of $[284, 408]$ and is positive in all 200 resamples.

\paragraph{Partially evaluated methods}
TabLLM was run on 34 classification datasets only. Battles are formed only between methods run on the same split, so TabLLM is rated from its 5{,}845 battles on those datasets. Its wider interval reflects the smaller number of battles.

\begin{table}[h]
    \caption{\textbf{Elo ratings of all methods}, as plotted in \autoref{fig:elo}, at $n_\mathrm{train}=1{,}024$. The interval columns give the distance from the rating to the upper and lower bounds of its 95\% bootstrap interval. Ratings are shifted so that their mean over the 36 methods is 1000. Every method is rated on all 100 datasets, except TabLLM, which ran on 34 classification datasets only. LLM enc.\ is Llama-3.1-8B applied to each cell in isolation; TF-IDF is followed by an SVD; \emph{CASE-like} is CASE with Gemma-3-4B (\autoref{app:baselines}).}
    \label{tab:elo}
    \centering
    \footnotesize
    \setlength{\tabcolsep}{6pt}
    \begin{tabular}{lrrr}
        \toprule
        \textbf{Method} & \textbf{Elo} & \multicolumn{2}{c}{\textbf{95\% interval}} \\
        \midrule
        TILE (Causilo) & 1500 & $+76$ & $-50$ \\
        TILE (TabPFN-3.5) & 1441 & $+59$ & $-42$ \\
        TILE (EXAONE-Tabular) & 1408 & $+63$ & $-44$ \\
        TILE (TabPFN-2.5) & 1342 & $+47$ & $-44$ \\
        TILE (TabICLv2) & 1315 & $+47$ & $-33$ \\
        TILE (TabPFN-3) & 1301 & $+46$ & $-33$ \\
        \addlinespace[2pt]
        CASE-like + TabPFN-3.5 & 1084 & $+48$ & $-51$ \\
        CASE-like + Causilo & 1053 & $+45$ & $-37$ \\
        CASE-like + EXAONE-Tabular & 1037 & $+41$ & $-33$ \\
        CASE-like + TabPFN-3 & 911 & $+44$ & $-45$ \\
        CASE-like + TabPFN-2.5 & 876 & $+46$ & $-48$ \\
        CASE-like + TabICLv2 & 864 & $+42$ & $-50$ \\
        \addlinespace[2pt]
        LLM enc.\ + EXAONE-Tabular & 1158 & $+43$ & $-41$ \\
        LLM enc.\ + Causilo & 1124 & $+43$ & $-41$ \\
        LLM enc.\ + TabPFN-3.5 & 1107 & $+45$ & $-44$ \\
        LLM enc.\ + TabPFN-2.5 & 1024 & $+32$ & $-32$ \\
        LLM enc.\ + TabICLv2 & 1008 & $+37$ & $-40$ \\
        LLM enc.\ + TabPFN-3 & 992 & $+31$ & $-32$ \\
        LLM enc.\ + XGBoost & 848 & $+25$ & $-27$ \\
        LLM enc.\ + RealMLP & 815 & $+30$ & $-39$ \\
        LLM enc.\ + CatBoost & 770 & $+32$ & $-37$ \\
        LLM enc.\ + Ridge & 744 & $+42$ & $-50$ \\
        \addlinespace[2pt]
        TF-IDF + TabPFN-3.5 & 1107 & $+30$ & $-37$ \\
        TF-IDF + EXAONE-Tabular & 1090 & $+30$ & $-31$ \\
        TF-IDF + Causilo & 1070 & $+30$ & $-34$ \\
        TF-IDF + TabPFN-2.5 & 991 & $+29$ & $-33$ \\
        TF-IDF + TabICLv2 & 950 & $+27$ & $-34$ \\
        TF-IDF + TabPFN-3 & 929 & $+34$ & $-35$ \\
        TF-IDF + XGBoost & 780 & $+27$ & $-33$ \\
        TF-IDF + CatBoost & 761 & $+30$ & $-33$ \\
        TF-IDF + RealMLP & 743 & $+32$ & $-35$ \\
        TF-IDF + Ridge & 574 & $+47$ & $-56$ \\
        \addlinespace[2pt]
        ContextTab & 914 & $+35$ & $-34$ \\
        TabLLM (T0-3B) & 903 & $+95$ & $-82$ \\
        TabSTAR & 816 & $+46$ & $-45$ \\
        CatBoost & 652 & $+42$ & $-49$ \\
        \bottomrule
    \end{tabular}
\end{table}

\clearpage
\section{The TILE pipeline, end to end}\label{app:pipeline}

Putting the above together, TILE maps a row $\mathbf{x}$ to a prediction as follows; every statistic (standardization, PCA, target encoding) is fitted on the training rows only.
\begin{enumerate}[noitemsep,topsep=2pt,parsep=0pt,partopsep=0pt,leftmargin=3ex]
    \item \textbf{Serialize.} $s(\mathbf{x})$ concatenates ``The <column name> is <cell value>'' over the $M$ columns; tokenizing gives $\mathbf{t} \in \mathcal{V}^{L}$ and the token span of each column.
    \item \textbf{Encode.} A forward pass through the LLM truncated at layer $\ell$ gives $H = \mathrm{LLM}_{1:\ell}(\mathbf{t}) \in \mathbb{R}^{L \times d}$.
    \item \textbf{Pool by column.} Equation \ref{eq:columnar_pooling}, then standardization, gives $\tilde{c}_j$ for every $j \in \mathcal{S}$.
    \item \textbf{Project by column.} Equation \ref{eq:row_representation} gives $\mathbf{u}$, with $k$ set by the learner's feature-capacity window.
    \item \textbf{Append the by-pass features.} $\tilde{\mathbf{x}} = [\mathbf{u}; \mathbf{x}_{\mathrm{num}}; \mathbf{x}_{\mathrm{cat}}; \mathbf{x}_{\mathrm{te}}]$, none of which passes through the PCA.
    \item \textbf{Predict.} $\hat{y} = \mathrm{PFN}\big(\tilde{\mathbf{x}}_{\mathrm{query}} \mid \{(\tilde{\mathbf{x}}_i, y_i)\}_{i=1}^{n}\big)$, in one forward pass over the training and test rows.
\end{enumerate}

\paragraph{Embedding extraction} Each LLM is truncated after the block of its extraction layer (\autoref{sec:probing}), and its hidden states are read there, without the final normalization layer that precedes the output head: Llama-3.1-8B, for instance, keeps its first 14 of 32 blocks. The whole serialized row is tokenized at once, and the token span of each cell is located on that tokenization; a token straddling two cells is mean-pooled into both. Missing values and blank strings are written ``NaN'' in the serialization. The model runs in bfloat16 on batches of up to $65{,}536$ tokens, and the embeddings are stored in float16.

\clearpage
\section{Target encoding of the categorical string columns}\label{app:target_encoding}

Alongside its LLM embeddings, TILE passes every suitable string column to the tabular learner as a target encoding \citep{micci2001preprocessing}. This section specifies that step; \autoref{sec:multimodal} motivates it.

\paragraph{Encoding} For a string column taking the value $c$, classification with $K$ target classes encodes $c$ by the class proportions of the training rows carrying it, $\hat p_k(c) = n_{c,k}/n_c$: one column per target class, or the single $\hat p_1(c)$ for a binary target, since the proportions sum to one. Similarly, regression uses the category mean $n_c^{-1}\sum_{i : c_i = c} y_i$, and adds a single column. Missing entries form their own category.

\paragraph{Out-of-fold estimation} Training rows are encoded from five folds so that a row never contributes to its own value (the rows of one fold use the statistics of the other four). Without that, a high-cardinality column would encode every singleton category to its own label, and the learner would read the target off the encoding. Test rows are encoded from the whole training split. Categories seen at test time but absent from the training rows fall back to the prior: the global target class proportions for classification or the global target mean $\bar y$ for regression.

\paragraph{Which columns are encoded} A string column is target-encoded when more than $10\%$ of its non-missing entries hold a value that occurs at least twice. Below that the column is almost entirely singletons and would encode to a near-constant.

\paragraph{Integration in the pipeline} The encoded columns are appended as numerical features to the PCA-reduced embedding and to the table's own numerical and low-cardinality categorical columns, and are not themselves passed through the PCA. This path is independent of the categorical by-pass: a column with fewer than 50 distinct values also reaches the learner as a native categorical feature, whereas a high-cardinality column reaches it only through its embedding and its target encoding.

\clearpage
\section{Baselines and their implementation}\label{app:baselines}

We compare against three families of string handling. (i) Learners with internal string handling: CatBoost, ContextTab and TabSTAR. (ii) TF-IDF followed by an SVD, and (iii) the LLM encoder (Llama-3.1-8B applied to the cell content), each combined with seven tabular learners: Ridge, XGBoost, CatBoost, RealMLP, TabPFN-2.5, TabPFN-3 and TabICLv2. This gives 17 baselines, all of which ran on the full benchmark (65 regression and 35 classification datasets) with 5 seeds.

These 17 baselines, together with the three default TILE variants (using Llama-3.1-8B as LLM backbone), form the reference pool used to normalize the scores reported throughout the paper (\autoref{sec:normalization}). We additionally report TabLLM \citep{hegselmann2023tabllm}, a fine-tuned LLM that predicts by generating text. Since it is a classifier and runs out of memory on one classification dataset, it covers only part of the benchmark and does not enter the reference pool. The concurrent CASE \citep{schindler2026enhancing} is reported outside the pool as well, and so are three recent table foundation models: TabPFN-3.5 \citep{jager2026tabpfn}, EXAONE-Tabular \citep{eo2026exaone} and Causilo \citep{cho2026causilo}.

Every baseline sees exactly the splits TILE sees: the same five seeds and the same training and test rows (\autoref{sec:normalization}), stratified for classification, and scored with the same metric code, ROC~AUC (one-vs-rest, weighted) for classification and $R^2$ for regression.

\paragraph{Learners with internal string handling} \textbf{CatBoost} (v1.2.10) is run on the raw table with the string columns declared as \texttt{cat\_features}: columns are typed with a \texttt{skrub} \texttt{TableVectorizer} (\texttt{cardinality\_threshold=0}, \texttt{high\_cardinality="passthrough"}) so that no column is one-hot encoded and every non-numeric column is passed to CatBoost as a categorical feature, with missing entries mapped to the string \texttt{"nan"}.
\textbf{TabSTAR} \citep{arazi2025tabstar} is the official \texttt{tabstar} package (v1.1.15) with its default pre-trained checkpoint; numeric columns are kept numeric, datetimes are decomposed with \texttt{skrub}'s \texttt{DatetimeEncoder}, and all remaining columns are cast to text so the model sees them through its text encoder.
\textbf{ContextTab} \citep{spinaci2025contexttab} is the official SAP release \texttt{sap-rpt-oss} (v1.0.1) with default settings, on tables passed through \texttt{skrub}'s \texttt{Cleaner}.

\paragraph{TF-IDF string representations} Every non-numeric column is embedded independently with \texttt{skrub}'s \texttt{StringEncoder} (v0.7.0), a character $n$-gram TF-IDF followed by a truncated SVD. The SVD width is 30 components per column, reduced to $\lfloor 500 / n_{\mathrm{str}} \rfloor$ when the table has more than 17 string columns, to keep the representation of wide tables tractable. The resulting matrix is concatenated with the numeric columns and handed to each of the seven tabular learners listed below.

\paragraph{LLM encoder} This baseline is \texttt{skrub}'s \texttt{LLMEncoder} (formerly \texttt{TextEncoder}). Each string cell is embedded on its own by Llama-3.1-8B (\texttt{meta-llama/Llama-3.1-8B}) loaded through \texttt{sentence-transformers} (v5.2.0); embeddings are computed once per unique cell value and broadcast back to the table. Each column's $4096$-dimensional block is then reduced to $30$ dimensions by a column-wise PCA fitted on the training rows, concatenated across columns and with the numeric columns, and fed to the same seven learners. This baseline is the natural ablation of TILE: the same language model, but applied at full depth to isolated cells without inter-column attention.

\paragraph{Downstream tabular learners} Both representation families are paired with \textbf{Ridge} (\texttt{scikit-learn} v1.7.2, \texttt{RidgeCV}/\texttt{RidgeClassifierCV} with $\alpha \in \{10^{-4},\dots,10^{4}\}$, numeric columns mean-imputed and power-transformed), \textbf{XGBoost} \citep{chen2016xgboost} (v3.2.0), \textbf{CatBoost} \citep{prokhorenkova2018catboost} (v1.2.10), \textbf{RealMLP} \citep{holzmuller2024better} (\texttt{pytabkit} v1.7.3), \textbf{TabPFN-2.5} and \textbf{TabPFN-3} \citep{grinsztajn2025tabpfn, grinsztajn2026tabpfn} (\texttt{tabpfn} v8.0.6, default checkpoint of each version, \texttt{ignore\_pretraining\_limits} enabled), and \textbf{TabICLv2} \citep{qu2026tabiclv2} (\texttt{tabicl} v2.1.1, default checkpoint). Unless stated otherwise, learners are run with their default hyper-parameters, which is the recommended usage of the table foundation models. Two exceptions: XGBoost is tuned by a 100-iteration random search over the grid of \autoref{tab:xgb-grid}, evaluated by $5$-fold cross-validation repeated $5$ times on the training rows; and RealMLP uses its own built-in hyper-parameter optimization (\texttt{RealMLP\_HPO}, 50 search steps, internal $5$-fold cross-validation, validation metric \texttt{1-auc\_ovr} for classification).

\paragraph{Additional table foundation models} We use \textbf{TabPFN-3.5} \citep{jager2026tabpfn} through \texttt{tabpfn} v9.0.0 with the \texttt{v3.5-20260909} checkpoint, \textbf{EXAONE-Tabular} \citep{eo2026exaone} through \texttt{exaonetabular} v1.0.0 with the v1 default checkpoints (Hugging Face revision \texttt{093ad1c}), and \textbf{Causilo} \citep{cho2026causilo} through \texttt{causilo} v1.0.2 with its release checkpoints (revision \texttt{94f2bd9}). All three run with their default inference settings (8 ensemble members for EXAONE-Tabular and Causilo) and no hyper-parameter tuning.

\begin{table}[h]
    \caption{\textbf{Search space of the tuned XGBoost baseline.} Sampled by a 100-iteration random search, scored by $5$-fold cross-validation repeated $5$ times on the training rows.}
    \label{tab:xgb-grid}
    \centering
    \small
    \begin{tabular}{llr}
        \toprule
        \textbf{Hyperparameter} & \textbf{Distribution} & \textbf{Range} \\
        \midrule
        \texttt{n\_estimators} & Integer & [50, 1000] \\
        \texttt{max\_depth} & Integer & [2, 6] \\
        \texttt{min\_child\_weight} & Log-uniform & [1, 100] \\
        \texttt{subsample} & Uniform & [0.5, 1.0] \\
        \texttt{learning\_rate} & Log-uniform & [$10^{-5}$, 1] \\
        \texttt{colsample\_bylevel} & Uniform & [0.5, 1.0] \\
        \texttt{colsample\_bytree} & Uniform & [0.5, 1.0] \\
        \texttt{gamma} & Log-uniform & [$10^{-8}$, 7] \\
        \texttt{reg\_lambda} & Log-uniform & [1, 4] \\
        \texttt{alpha} & Log-uniform & [$10^{-8}$, 100] \\
        \bottomrule
    \end{tabular}
\end{table}

\paragraph{TabLLM} TabLLM \citep{hegselmann2023tabllm} serializes each row as a sentence (\emph{Text} template: \texttt{"The <column> is <value>."}) and fine-tunes a T0 model \citep{sanh2022multitask} with (IA)$^3$ following the t-few recipe \citep{liu2022few}. We run the original TabLLM and t-few code at pinned commits, with the configuration behind the paper's main results---\emph{Text} serialization, (IA)$^3$ initialized from the 100k-step multitask pre-trained checkpoint, 30 epochs, batch size 4, learning rate $3\cdot10^{-3}$, length-normalized answer-choice scoring, inputs truncated at 1024 tokens---with one substitution: we use T0-3B rather than T0-11B, the variant TabLLM reports in its appendix and that reaches similar performance, because T0-11B runs out of memory at $n_{\mathrm{train}}=1{,}024$ samples (and otherwise would cost on the order of $10^3$ GPU-hours). Two further deviations are required for comparability: the model is trained on the benchmark's fixed stratified splits (\texttt{num\_shot=all} on exactly the $n_{\mathrm{train}}$ rows every other baseline sees) instead of TabLLM's own $80/20$ re-split and class-balanced sampling with replacement, and the predictions are scored downstream with the same metric code as every other baseline. Since TabLLM is a classifier, it is evaluated on the classification part of the benchmark only.

\paragraph{CASE-like} We run CASE \citep{schindler2026enhancing} with the authors' official implementation (commit \texttt{64a0a79}) and its default configuration, but without the table-tuned TabGemma checkpoints of the paper, that have not been released. We use the fallback backbone recommended in the official repository, \texttt{google/gemma-3-4b-pt}, in \texttt{bfloat16}. All other settings are the library defaults: the KV cache is prefilled with 128 training rows and their targets under a 16k-token budget, with at most 50 tokens per cell; each row is embedded as the last-layer hidden state at its final token; and the row embeddings are reduced to 32 components by PCA and appended to the raw features, which the downstream learner preprocesses itself. Our only deviation is the attention implementation: PyTorch's SDPA rather than FlashAttention-2, which computes the same attention without a compiled dependency. CASE is fit per split, with the context rows and the PCA taken from the training rows only. The resulting features are shared by TabICLv2, TabPFN-2.5 and TabPFN-3, run at their default settings on the same splits as every other baseline. Reported preprocessing time covers embedding the training and test rows and excludes loading the backbone, which is done once per job.

\clearpage
\section{Additional experimental results}\label{app:extra_results}

\subsection{The improvability metric}\label{app:improvability}
The cost analyses report the \emph{improvability} of \citet{erickson2025tabarena}, the fraction of a method's error ($1-R^2$ for regression, $1-{}$AUC for classification) that would be removed by switching to the best available method on each dataset, averaged over datasets and seeds. A method that is best on every dataset has an improvability of $0\%$.

\subsection{The feature-capacity window of each learner}\label{sec:PCA}

\begin{figure}[h]
    \centering
    \includegraphics[width=\linewidth]{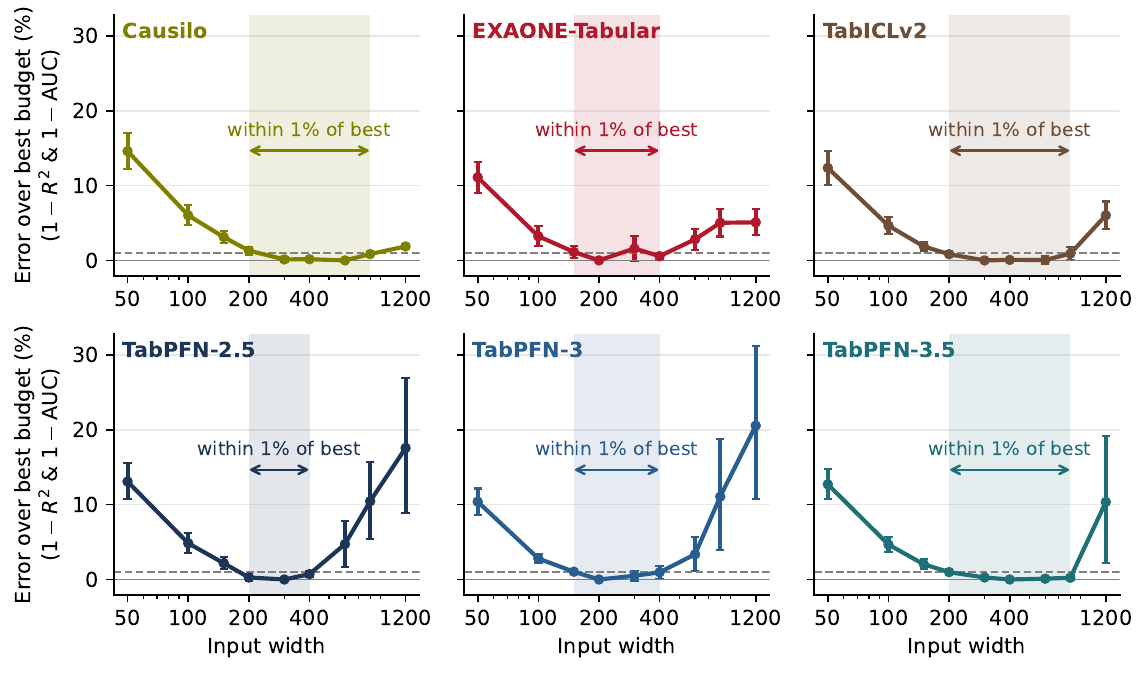}
    \caption{\textbf{TFMs need dimensionality reduction.} Error relative to the best budget, as a function of the learner's total input width (reduced embeddings plus by-pass features). Dots are geometric means over the 27 probing tables ($\pm$ one standard error). The shaded areas where the performance stays within $1\%$ of the best budget are the feature-capacity windows used in TILE (\autoref{sec:llm_route}).}
    \label{fig:pca-8b}
\end{figure}

How many features can a table foundation model actually digest? The number of features used to pre-train PFNs is usually much lower than the number of dimensions of an LLM embedding \citep{tajjar2026pretraining, wang2026limix2m}, and being transformers they likely inherit the shortcomings of attention with large contexts \citep{liu2024lost}. Too narrow an input discards information, while too wide a one hands the learner more features than it handles comfortably (\autoref{fig:pca-8b}). A learner's \emph{feature-capacity window} $[a, b]$ bounds the total width of its input: the PCA-reduced embeddings plus the features that by-pass them (numerical, target-encoded and native categorical columns). Each string column keeps $30$ components of its own PCA by default; whatever width the by-pass features leave in the window, these counts are rescaled to fit it, with at least one component per column (\autoref{sec:llm_route}).

\paragraph{Selection procedure} The window is selected automatically on the 27 probing tables (\autoref{sec:probing_datasets}). We run the learner on each table with its total input width fixed to each budget of a grid from $50$ to $1{,}200$ features, using the same per-column PCA as TILE. On each table, we divide the error at every budget ($1-R^2$, with $R^2$ clipped at 0, or $1-{}$AUC) by the error at that table's best budget. This relative error scales the tolerance with the difficulty of the table. We then average these relative errors over the tables with a geometric mean, and call the budget with the lowest average the best one. The window is the contiguous range of budgets around it whose average relative error stays within $1\%$ of the best, allowing for one standard error: a budget is kept unless it is clearly worse than the best one.
This gives $[200, 400]$ for TabPFN-2.5, $[200, 800]$ for TabICLv2, $[150, 400]$ for TabPFN-3, $[200, 800]$ for TabPFN-3.5, $[150, 400]$ for EXAONE-Tabular and $[200, 800]$ for Causilo.

\paragraph{An ideal learner would not need the window} A larger budget means a milder dimensionality reduction, and thus more of the information carried by the embeddings. A learner able to exploit all of its input would therefore see its error decrease steadily with the budget, and the curves of \autoref{fig:pca-8b} would be non-increasing. Instead, the error of every learner rises again beyond its window, highlighting that current TFMs do not handle that many features well, and need the dimensionality reduction to make use of the embeddings.

\subsection{Cardinality threshold for categorical columns}\label{app:cardinality}

\begin{figure}[h]
    \centering
    \includegraphics[width=\linewidth]{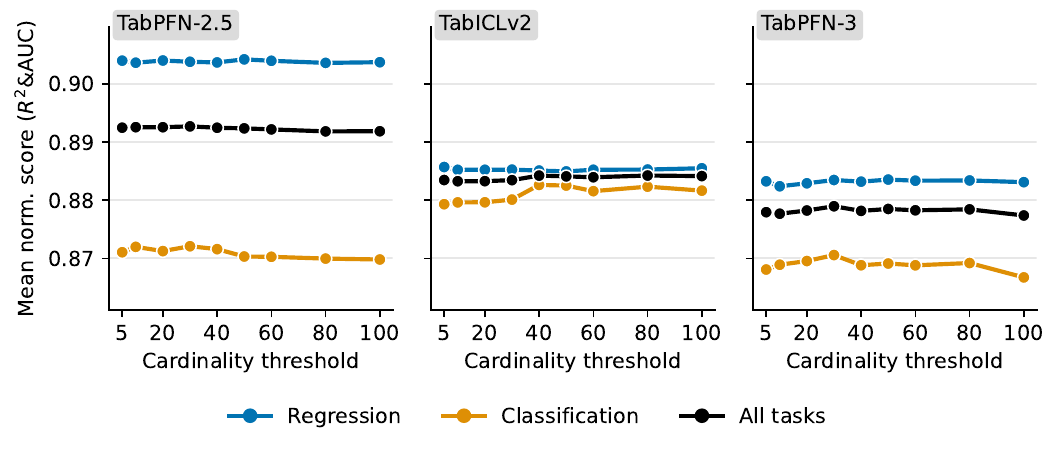}
    \caption{\textbf{The categorical cardinality threshold has little effect.} Averaged normalized score of TILE on the 27 probing tables as a function of the cardinality threshold below which a string column is also passed directly to the tabular learner, for TabPFN-2.5, TabICLv2 and TabPFN-3. The curves are mostly flat. We use 50 in all other experiments.}
    \label{fig:cardinality}
\end{figure}

TILE passes low-cardinality string columns to the tabular learner as native categorical features in addition to target encoding and serializing them. The cardinality threshold decides if a string column by-passes the LLM or not. It is selected on the 27 probing tables only (\autoref{sec:probing_datasets}). \autoref{fig:cardinality} sweeps the threshold from 5 to 100 on these tables, for TabPFN-2.5, TabICLv2 and TabPFN-3. The curves are mostly flat for all three learners and both problem types, so the threshold matters little. We use 50 by default, but other values would serve as well.

\subsection{Choosing the pooling strategy}\label{app:pooling}

\begin{figure}[h]
    \centering
    \includegraphics[width=0.7\linewidth]{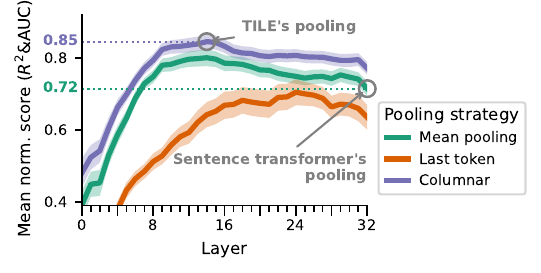}
    \caption{\textbf{Columnar pooling gives the best representations.} Performance of three pooling strategies across the layers of Llama-3.1-8B, scored by Ridge on the embeddings alone (no by-pass features), on the 27 probing datasets of \autoref{sec:probing_datasets}. Ridge is used without PCA, so columnar pooling gives it a higher-dimensional input than the other two strategies. Scores are normalized over the strategies shown here, so they are not comparable with those of the other figures.}
    \label{fig:pooling-8b}
\end{figure}

\autoref{fig:pooling-8b} compares three ways of pooling the token embeddings of Llama-3.1-8B into a row representation, at every layer: mean pooling over the whole sequence, the embedding of the last token, and columnar pooling. As in \autoref{sec:probing}, the representations are scored with Ridge on the embeddings alone, without the by-pass features, on the 27 probing datasets of \autoref{sec:probing_datasets}. Columnar pooling is ahead at every layer, and at layer 14 it reaches $0.85$, against $0.72$ for mean pooling of the last layer, the default of the sentence-transformers library; we therefore use it in TILE. This comparison favors columnar pooling in one respect: Ridge is applied without PCA, so columnar pooling hands it one embedding per column, an input $M$ times wider than the single vector of mean or last-token pooling, and part of its advantage may come from that extra dimensionality.

\FloatBarrier

\clearpage
\section{Compute and memory costs}\label{app:compute_resources}

\subsection{Hardware and total compute}

All experiments, for TILE and all baselines, were conducted on a high-performance computing cluster, on NVIDIA A100, A40 and H100 GPUs (\autoref{app:pareto_tasks}), but TILE can be run on a smaller GPU. TILE with Llama-3.1-8B runs on a single GPU and 15 CPU cores: its extraction peaks at $17.7$ GB, or about $8.7$ GB with smaller batches (\autoref{app:memory}). 

In total, this project consumed approximately 15,200 GPU compute hours. This total consists of an estimated 6,500 GPU hours dedicated to developing our method, running ablation studies, and conducting exploratory research, alongside 8,700 GPU hours for the evaluation and hyperparameter tuning of the baseline models.

\subsection{Runtime by task}\label{app:pareto_tasks}

\begin{figure}[h]
    \centering
    \includegraphics[width=\linewidth]{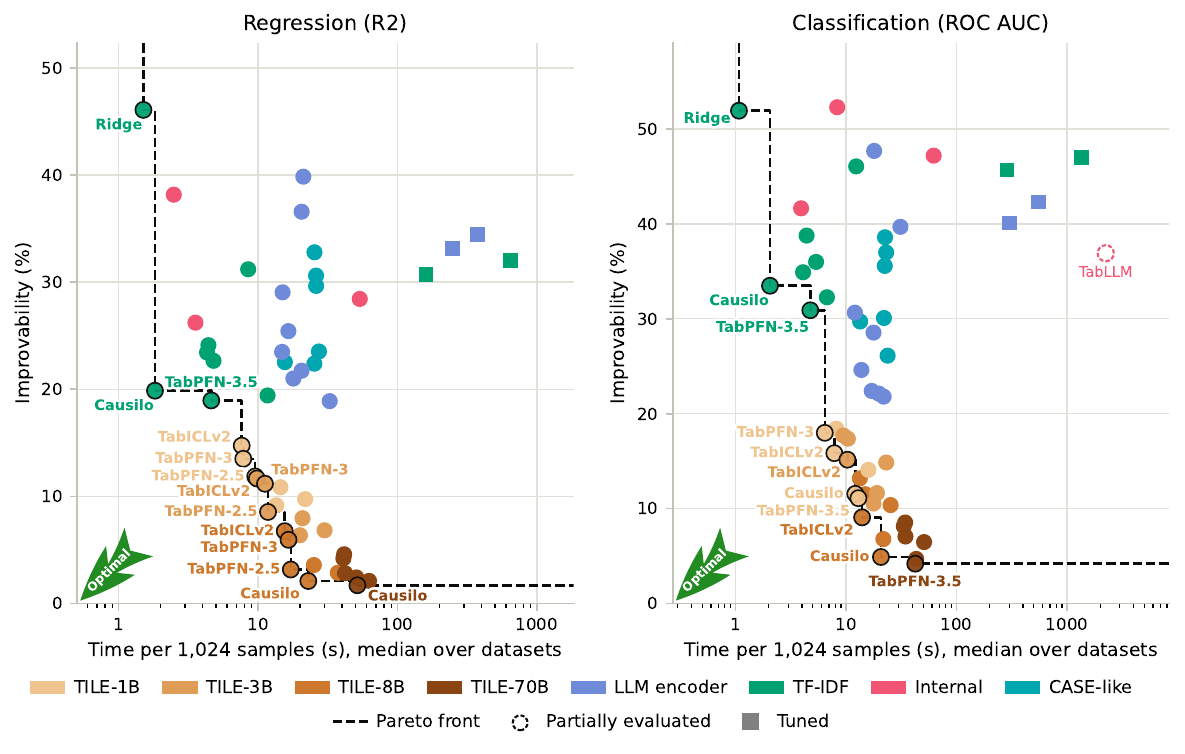}
    \caption{\textbf{TILE holds the accurate end of the Pareto front.} Improvability against median time per dataset at $n_\mathrm{train}=1{,}024$, split into regression (left) and classification (right). TILE is run with the four Llama backbones, from TILE-1B (Llama-3.2-1B, lightest brown) to TILE-70B (Llama-3.1-70B, darkest). The dashed staircase is the Pareto front, and only Pareto-optimal pipelines are labeled, with their learner. Colors indicate the string encoding; squares mark tuned baselines. TabLLM supports classification only, and is evaluated on the 34 datasets it completes.}
    \label{fig:pareto_time_tasks}
\end{figure}

\autoref{fig:pareto_time_tasks} plots improvability against median runtime per dataset, separately for each problem type, with TILE run on the four Llama backbones of \autoref{sec:serialization}. TF-IDF pipelines hold the cheap end of the Pareto front, with Ridge, then Causilo and TabPFN-3.5, and TILE holds all the rest of it. A larger backbone buys accuracy with time. The median time per dataset of the TILE variants grows with the backbone: $7$ to $22$\,s with Llama-3.2-1B, $10$ to $30$\,s with Llama-3.2-3B, $15$ to $37$\,s with Llama-3.1-8B, and $40$ to $63$\,s with Llama-3.1-70B. Even with Llama-3.2-1B, every TILE variant is more accurate than every baseline, in less time than most cell-wise LLM-encoder and CASE-like pipelines ($14$ to $37$\,s). The best baseline, an LLM-encoder pipeline, sits at $19\%$ improvability in regression and $22\%$ in classification, against $2.1\%$ and $4.9\%$ for TILE with Llama-3.1-8B and Causilo, and $1.7\%$ and $4.2\%$ for the best TILE with Llama-3.1-70B. Of the error that separates the best existing pipeline from the per-dataset best, TILE with Llama-3.1-8B thus removes nearly nine tenths in regression and more than three quarters in classification.

We ran TILE pipelines on A100 or A40 GPUs, except the embedding extraction of TILE with Llama-3.1-70B, which ran on one H100; this faster hardware slightly favors the runtimes of that backbone. Other baselines also ran on A100 or A40 GPUs, except CASE-like for which some datasets produced out-of-memory errors and were run on H100 GPUs instead.

\subsection{Peak GPU memory}\label{app:memory}

\begin{figure}[h]
    \centering
    \includegraphics[width=\linewidth]{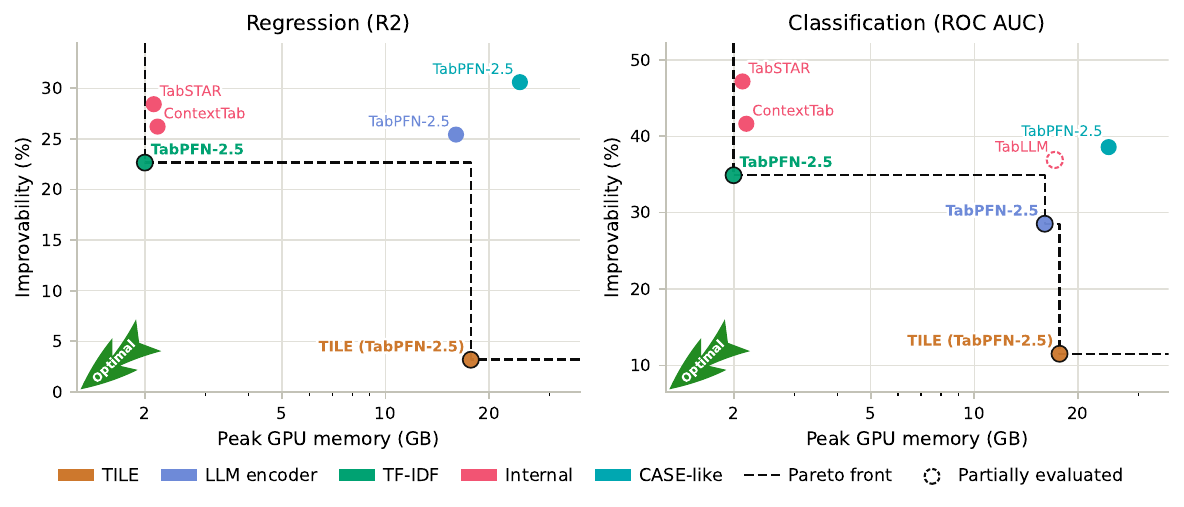}
    \caption{\textbf{TILE can trade memory for speed.} Improvability (\autoref{app:improvability}, lower is better) against the peak GPU memory a method needs on $1{,}024$ train rows of the \texttt{chocolate\_bar\_ratings} dataset of CARTE \citep{kim2024carte}, taken as the larger of its two stages: building the representation, then fitting the learner and predicting (CASE-like keeps both loaded, so its footprint covers both). As in \autoref{fig:pareto_time_tasks}, both axes are costs and the dashed staircase is the Pareto front. TILE extracts embeddings in batches of up to $65{,}536$ tokens; with batches of 16 rows, its peak drops to about $8.7$ GB. We report it for TILE, for TF-IDF, the LLM encoder and CASE-like combined with the same learner (TabPFN-2.5), and for the three end-to-end models, which read the raw table with a single network. TabLLM supports classification only, its improvability is computed on the 34 datasets it completes.}
    \label{fig:pareto_memory}
\end{figure}

\autoref{fig:pareto_memory} plots improvability against the peak GPU memory of each method, measured on the same $1{,}024$ training rows of one dataset (CARTE \texttt{chocolate\_bar\_ratings}). For two-stage methods, the LLM and the tabular learner run one after the other, so the footprint is the larger of the two stages. CASE-like keeps its backbone and its learner loaded at the same time, and its footprint covers both. TILE peaks at $17.7$ GB, CASE-like at $24.6$ GB, and the LLM encoder at $16.0$ GB, while TILE has the lowest improvability in both regression ($3.2\%$ vs.\ $25.4\%$ for the LLM encoder) and classification ($11.5\%$ vs.\ $28.5\%$). TILE's peak comes from its embedding extraction and mostly reflects its batch size, not its model. TILE reads its embeddings at layer 14 of Llama-3.1-8B and therefore loads only the first 14 of its 32 blocks, $6.7$ GB of weights in bfloat16, about half the $14$ GB of the full model that the LLM encoder loads. The rest of TILE's peak is activations, because the extraction groups rows into batches of up to $65{,}536$ tokens to make it faster. With a batch of 16 rows, comparable to the LLM encoder's, TILE's extraction peaks at about $8.7$ GB, roughly half the memory of the LLM encoder and a third of that of CASE-like. Memory can therefore be traded for extraction speed through the batch size.

TF-IDF with TabPFN-2.5 holds the frugal end of the Pareto front at $2.0$ GB, and ContextTab and TabSTAR sit just beside it at $2.1$--$2.2$ GB while being markedly less accurate. In classification, the LLM encoder also reaches the front, being slightly lighter than TILE at its default batch size. TabLLM needs $17.1$ GB, about as much as TILE, to generate its predictions as text, for a much higher improvability. CASE-like is the heaviest method and, at $30.6\%$ improvability in regression and $38.6\%$ in classification, it sits in the dominated top-right corner of both panels.

\clearpage
\section{Raw results}\label{app:raw_results}

\begin{table}[t]
    \caption{\textbf{Raw scores at every training-set size.} Unnormalized counterpart of \autoref{tab:large_datasets}: $R^2$ for regression and ROC-AUC for classification, multiplied by 100 and aggregated over the datasets large enough for each training size; their number in parentheses. Because the set of datasets changes with the training size, compare scores within a column rather than across. LLM enc.\ is Llama-3.1-8B applied to each cell content in isolation, the same LLM as TILE; TF-IDF is followed by an SVD; \emph{CASE-like} is CASE with Gemma-3-4B (\autoref{app:baselines}). Best result in bold, second best underlined.}
    \label{tab:raw_results}
    \centering
    \footnotesize
    \setlength{\tabcolsep}{4.5pt}
    \begin{tabular}{lcccccc}
        \toprule
        & \multicolumn{3}{c}{\textbf{Regression ($R^2$)}} & \multicolumn{3}{c}{\textbf{Classification (AUC)}} \\
        \cmidrule(lr){2-4} \cmidrule(lr){5-7}
        \textbf{Method} & \textbf{1k (65)} & \textbf{10k (38)} & \textbf{50k (7)} & \textbf{1k (35)} & \textbf{10k (10)} & \textbf{50k (2)} \\
        \midrule
        TILE (TabPFN-2.5) & 71.03 & 76.95 & 82.66 & 89.39 & \underline{90.54} & 82.63 \\
        TILE (TabICLv2) & 70.27 & 76.52 & 82.83 & 89.56 & 90.18 & 82.76 \\
        TILE (TabPFN-3) & 70.28 & 76.48 & 82.48 & 89.31 & 90.43 & 82.69 \\
        TILE (TabPFN-3.5) & \underline{71.12} & \underline{77.01} & \underline{83.39} & \underline{89.67} & 90.47 & \textbf{83.21} \\
        TILE (Causilo) & \textbf{71.56} & \textbf{77.77} & \textbf{84.04} & \textbf{89.77} & \textbf{90.58} & \underline{83.15} \\
        \addlinespace[2pt]
        CASE-like + TabPFN-2.5 & 58.36 & 61.78 & 53.90 & 86.83 & 89.28 & 78.43 \\
        CASE-like + TabICLv2 & 57.09 & 62.13 & 55.95 & 87.31 & 89.38 & 79.21 \\
        CASE-like + TabPFN-3 & 58.27 & 62.49 & 55.32 & 86.80 & 89.45 & 79.45 \\
        CASE-like + TabPFN-3.5 & 61.82 & 67.18 & 59.47 & 88.19 & 90.21 & 81.22 \\
        CASE-like + Causilo & 63.25 & 70.86 & 76.30 & 87.99 & 89.75 & 80.79 \\
        \addlinespace[2pt]
        ContextTab & 63.10 & 70.99 & 74.54 & 87.09 & 89.36 & 79.71 \\
        TabSTAR & 61.93 & 71.05 & 79.42 & 86.11 & 88.98 & 79.48 \\
        CatBoost & 55.70 & 65.69 & 70.88 & 83.24 & 88.98 & 80.76 \\
        \addlinespace[2pt]
        LLM enc.\ + TabPFN-2.5 & 62.94 & 72.39 & 79.88 & 88.33 & 89.97 & 80.01 \\
        LLM enc.\ + TabICLv2 & 60.64 & 72.71 & 79.96 & 88.55 & 89.98 & 81.27 \\
        LLM enc.\ + TabPFN-3 & 63.88 & 70.67 & 79.30 & 87.65 & 88.95 & 79.83 \\
        LLM enc.\ + TabPFN-3.5 & 64.02 & 73.60 & 79.96 & 88.55 & 89.93 & 80.73 \\
        LLM enc.\ + Causilo & 64.75 & 73.46 & 81.12 & 88.60 & 89.95 & 80.98 \\
        LLM enc.\ + RealMLP & 58.07 & 70.48 & 80.26 & 86.44 & 89.20 & 80.03 \\
        LLM enc.\ + XGBoost & 58.95 & 70.75 & 79.59 & 87.29 & 89.59 & 80.67 \\
        LLM enc.\ + CatBoost & 54.96 & 71.29 & 79.24 & 87.17 & 89.53 & 80.57 \\
        LLM enc.\ + Ridge & 56.91 & 63.92 & 74.05 & 86.15 & 87.51 & 78.14 \\
        \addlinespace[2pt]
        TF-IDF + TabPFN-2.5 & 63.95 & 71.73 & 78.74 & 87.34 & 89.33 & 79.80 \\
        TF-IDF + TabICLv2 & 62.97 & 71.11 & 79.24 & 87.32 & 89.27 & 80.45 \\
        TF-IDF + TabPFN-3 & 63.34 & 70.99 & 79.69 & 86.66 & 89.01 & 79.69 \\
        TF-IDF + TabPFN-3.5 & 65.37 & 74.03 & 81.26 & 87.87 & 89.89 & 81.59 \\
        TF-IDF + Causilo & 64.60 & 73.17 & 80.59 & 87.55 & 89.54 & 80.96 \\
        TF-IDF + RealMLP & 60.14 & 70.34 & 78.87 & 85.65 & 88.45 & 79.77 \\
        TF-IDF + XGBoost & 60.83 & 69.96 & 78.40 & 86.02 & 88.71 & 79.95 \\
        TF-IDF + CatBoost & 60.53 & 70.06 & 77.92 & 85.89 & 88.67 & 80.09 \\
        TF-IDF + Ridge & 48.81 & 32.44 & 44.51 & 84.26 & 75.84 & 75.72 \\
        \bottomrule
    \end{tabular}
\end{table}

Table~\ref{tab:raw_results} gives the raw scores behind \autoref{tab:large_datasets}, for every method and every training-set size. \autoref{tab:large_datasets_full} gives the normalized scores of every method, including the learners omitted from \autoref{tab:large_datasets} for space reasons.

\begin{table}[t]
    \caption{\textbf{Normalized scores at every training-set size, all learners.} Full version of \autoref{tab:large_datasets}. Normalized scores $\times 100$, aggregated over the datasets large enough for each size; counts in parentheses. Because the set of datasets varies, compare scores within columns only. LLM enc.\ is Llama-3.1-8B applied to each cell independently; TF-IDF is followed by SVD; \emph{CASE-like} is CASE with Gemma-3-4B (\autoref{app:baselines}). Best result in bold, second best underlined. \autoref{tab:raw_results} gives raw scores.}
    \label{tab:large_datasets_full}
    \centering
    \footnotesize
    \setlength{\tabcolsep}{6pt}
    \begin{tabular}{lcccccc}
        \toprule
        & \multicolumn{3}{c}{\textbf{Regression (normalized $R^2$)}} & \multicolumn{3}{c}{\textbf{Classification (normalized AUC)}} \\
        \cmidrule(lr){2-4} \cmidrule(lr){5-7}
        \textbf{Method} & \textbf{1,024 (65)} & \textbf{10k (38)} & \textbf{50k (7)} & \textbf{1,024 (35)} & \textbf{10k (10)} & \textbf{50k (2)} \\
        \midrule
        TILE (TabPFN-2.5) & 90.42 & 93.29 & 92.57 & 87.03 & 88.18 & 79.14 \\
        TILE (TabICLv2) & 88.50 & 92.23 & 92.61 & 88.25 & 85.66 & 80.68 \\
        TILE (TabPFN-3) & 88.36 & 91.94 & 91.89 & 86.91 & 87.32 & 80.22 \\
        TILE (TabPFN-3.5) & \underline{90.59} & \underline{93.66} & \underline{94.43} & \underline{89.59} & \underline{88.28} & \textbf{85.05} \\
        TILE (Causilo) & \textbf{91.76} & \textbf{94.99} & \textbf{95.88} & \textbf{89.97} & \textbf{88.33} & \underline{84.37} \\
        \addlinespace[2pt]
        CASE-like + TabPFN-2.5 & 61.68 & 64.75 & 49.14 & 67.23 & 77.31 & 34.91 \\
        CASE-like + TabICLv2 & 59.23 & 63.05 & 48.99 & 70.13 & 77.02 & 42.19 \\
        CASE-like + TabPFN-3 & 63.10 & 66.10 & 53.76 & 67.24 & 77.98 & 44.08 \\
        CASE-like + TabPFN-3.5 & 70.35 & 73.88 & 57.55 & 78.05 & 85.19 & 62.99 \\
        CASE-like + Causilo & 71.53 & 76.72 & 68.26 & 76.17 & 81.08 & 58.94 \\
        \addlinespace[2pt]
        ContextTab & 70.52 & 78.56 & 66.74 & 69.61 & 80.79 & 49.11 \\
        TabSTAR & 66.67 & 79.23 & 81.33 & 57.19 & 75.98 & 46.12 \\
        CatBoost & 52.16 & 64.19 & 51.96 & 40.82 & 75.93 & 61.80 \\
        \addlinespace[2pt]
        LLM enc. + TabPFN-2.5 & 72.59 & 82.88 & 85.34 & 79.07 & 83.50 & 51.91 \\
        LLM enc. + TabICLv2 & 68.46 & 83.33 & 85.77 & 80.16 & 83.75 & 66.19 \\
        LLM enc. + TabPFN-3 & 73.77 & 78.97 & 83.30 & 73.76 & 77.19 & 49.03 \\
        LLM enc. + TabPFN-3.5 & 75.99 & 85.72 & 85.93 & 80.24 & 83.79 & 60.10 \\
        LLM enc. + Causilo & 77.44 & 84.77 & 88.84 & 80.57 & 83.61 & 63.47 \\
        LLM enc. + RealMLP & 61.34 & 78.72 & 85.46 & 62.96 & 77.87 & 51.96 \\
        LLM enc. + XGBoost & 64.01 & 78.84 & 83.18 & 69.86 & 81.39 & 58.45 \\
        LLM enc. + CatBoost & 52.88 & 79.31 & 82.11 & 68.53 & 80.22 & 57.96 \\
        LLM enc. + Ridge & 56.10 & 62.80 & 66.19 & 53.33 & 60.33 & 32.17 \\
        \addlinespace[2pt]
        TF-IDF + TabPFN-2.5 & 72.58 & 80.11 & 79.62 & 72.21 & 78.65 & 50.54 \\
        TF-IDF + TabICLv2 & 70.74 & 79.00 & 79.92 & 71.42 & 78.11 & 57.15 \\
        TF-IDF + TabPFN-3 & 71.30 & 78.45 & 81.94 & 67.05 & 75.39 & 49.13 \\
        TF-IDF + TabPFN-3.5 & 76.22 & 85.80 & 86.79 & 76.08 & 82.83 & 69.18 \\
        TF-IDF + Causilo & 74.78 & 83.81 & 84.63 & 73.62 & 80.34 & 62.87 \\
        TF-IDF + RealMLP & 62.71 & 77.10 & 80.27 & 57.37 & 71.88 & 50.12 \\
        TF-IDF + XGBoost & 64.74 & 76.12 & 77.87 & 60.65 & 74.83 & 52.26 \\
        TF-IDF + CatBoost & 64.01 & 76.51 & 76.54 & 59.50 & 73.30 & 53.93 \\
        TF-IDF + Ridge & 39.64 & 17.19 & 12.41 & 38.43 & 15.16 & 12.14 \\
        \bottomrule
    \end{tabular}
\end{table}

\begin{table}[t]
    \caption{\textbf{Serialization ablation results aggregated over 100 datasets.} Scores are raw $R^2$/AUC multiplied by 100; best is in bold and second-best is underlined. These results correspond to the ablations presented in \autoref{fig:serialization}. The cell-wise LLM encoder could not be run with Llama-3.1-70B: the full-depth model weights exceed the memory of our GPUs. TILE runs with that backbone, as it only loads the layers up to the extraction point.}
    \label{tab:raw_serialization_results}
    \centering
    \small
    \setlength{\tabcolsep}{6pt}
    \renewcommand{\arraystretch}{1.05}
    \begin{tabular}{@{}llrr@{}}
        \toprule
        \textbf{Method} & \textbf{Base LLM} & \multicolumn{1}{c}{\textbf{Regression ($R^2$)}} & \multicolumn{1}{c}{\textbf{Classification (AUC)}} \\
        \midrule
        \multirow{7}{*}{TILE} 
        & Llama-3.1-70B & \textbf{72.38} & \textbf{89.83} \\
        & Llama-3.1-8B & \underline{71.03} & \underline{89.39} \\
        & Llama-3.2-3B & 69.65 & 89.16 \\
        & Llama-3.2-1B & 68.66 & 89.06 \\
        & Qwen3-8B-Base & 69.95 & 89.12 \\
        & Qwen3-4B-Base & 69.05 & 88.95 \\
        & Qwen3-0.6B-Base & 67.05 & 88.31 \\
        \midrule
        \multirow{7}{*}{\shortstack[l]{TILE w/o\\inter-col. attention}} 
        & Llama-3.1-70B & 67.50 & 89.23 \\
        & Llama-3.1-8B & 66.98 & 89.04 \\
        & Llama-3.2-3B & 65.45 & 88.88 \\
        & Llama-3.2-1B & 65.30 & 88.76 \\
        & Qwen3-8B-Base & 65.99 & 89.01 \\
        & Qwen3-4B-Base & 65.19 & 88.71 \\
        & Qwen3-0.6B-Base & 66.61 & 88.26 \\
        \midrule
        \multirow{7}{*}{\shortstack[l]{Base LLM on\\cell content\\(LLM encoder)}} 
        & Llama-3.1-70B & \textit{Out of memory} & \textit{Out of memory} \\
        & Llama-3.1-8B & 62.94 & 88.33 \\
        & Llama-3.2-3B & 61.53 & 88.33 \\
        & Llama-3.2-1B & 61.36 & 88.11 \\
        & Qwen3-8B-Base & 59.67 & 87.78 \\
        & Qwen3-4B-Base & 59.84 & 87.61 \\
        & Qwen3-0.6B-Base & 64.27 & 87.72 \\
        \bottomrule
    \end{tabular}
\end{table}

Table~\ref{tab:raw_serialization_results} reports how TILE's performance changes with the base LLM choice and with the ablations presented in \autoref{fig:serialization} (removing inter-column attention or using the base LLM directly on cell content, the LLM encoder).

\begin{table}[t]
    \caption{\textbf{Ablation study, raw scores.} The ablations of \autoref{tab:ablations}, with the raw $R^2$/AUC ($\times 100$) of the default architecture for each learner, and the raw difference obtained for each variation. Differences on classification are compressed here because ROC-AUC saturates near the top of its range; the normalized scores of \autoref{tab:ablations} place the two problem types on a comparable scale.}
    \label{tab:raw_ablations}
    \centering
    \small
    \setlength{\tabcolsep}{6pt}
    \renewcommand{\arraystretch}{1.05}
    \begin{tabular}{@{}lrrrrrr@{}}
        \toprule
        & \multicolumn{2}{c}{\textbf{TabPFN-2.5}} & \multicolumn{2}{c}{\textbf{TabICLv2}} & \multicolumn{2}{c}{\textbf{TabPFN-3}} \\
        \cmidrule(lr){2-3} \cmidrule(lr){4-5} \cmidrule(lr){6-7}
        \textbf{Model} & \textbf{AUC} & \textbf{$R^2$} & \textbf{AUC} & \textbf{$R^2$} & \textbf{AUC} & \textbf{$R^2$} \\
        \midrule
        $-$ numerical by-pass & -0.50 & -4.64 & -0.49 & -4.21 & -0.53 & -4.77 \\
        $-$ inter-column attention & -0.35 & -4.05 & -0.31 & -3.02 & -0.88 & -3.81 \\
        $-$ intermediate layer (last layer) & -0.01 & -1.21 & -0.02 & -1.14 & +0.01 & -1.06 \\
        $-$ feature-capacity window & -0.02 & -0.91 & -0.06 & -0.23 & +0.00 & -0.55 \\
        $-$ numbers in the serialization & +0.01 & -0.14 & -0.07 & -0.09 & -0.08 & -0.15 \\
        $-$ target-encoding by-pass & -0.03 & -0.08 & -0.11 & +0.06 & -0.08 & +0.08 \\
        $-$ standard scaling & -0.01 & -0.11 & -0.03 & -0.08 & -0.03 & -0.12 \\
        $-$ categorical by-pass & -0.02 & -0.03 & -0.01 & -0.00 & -0.03 & -0.04 \\
        \rowcolor{gray!20} TILE (default) & \textbf{89.39} & \textbf{71.03} & \textbf{89.56} & \textbf{70.27} & \textbf{89.31} & \textbf{70.28} \\
        $+$ numerical-cell embeddings & +0.00 & -0.08 & +0.04 & +0.10 & -0.04 & +0.07 \\
        $+$ one-hot instead of native cat. & -0.01 & -0.03 & -0.02 & -0.03 & -0.02 & -0.01 \\
        $+$ repeated serialization & -0.15 & -0.00 & -0.14 & -0.05 & -0.17 & -0.09 \\
        $+$ mirrored column order & -0.03 & -0.41 & -0.02 & -0.28 & -0.16 & -0.56 \\
        $+$ target name & -0.11 & -0.27 & -0.10 & -0.28 & -0.04 & -0.29 \\
        $+$ uninformative column names & -0.25 & -1.26 & -0.33 & -1.27 & -0.33 & -1.22 \\
        \midrule
        LLM encoder (\texttt{skrub}) & -1.06 & -8.09 & -1.02 & -9.63 & -1.66 & -6.40 \\
        LLM encoder $+$ by-pass features & -0.95 & -7.00 & -0.97 & -8.59 & -1.80 & -7.08 \\
        LLM encoder $+$ feature-capacity window & -1.03 & -7.67 & -0.97 & -9.41 & -2.00 & -8.12 \\
        \bottomrule
    \end{tabular}
\end{table}

\paragraph{Raw scores on STRABLE} \autoref{tab:strable_raw} gives the unnormalized scores behind \autoref{fig:strable}: the mean ROC-AUC over STRABLE's 32 classification datasets and the mean $R^2$ over its 76 regression datasets, for the TILE variants, with Llama-3.1-8B and Llama-3.1-70B, and the 63 STRABLE baselines. The baselines are run with STRABLE's protocol and implementations \citep{blayer2026strable}, not those of \autoref{app:baselines}. Most pair a string encoder with a tabular learner:
\begin{itemize}[leftmargin=*,itemsep=1pt,topsep=2pt]
    \item \textbf{String encoders:} TF-IDF; FastText \citep{bojanowski2017enriching}; the sentence encoders E5-small-v2 \citep{wang2022text}, All-MiniLM-L6-v2 \citep{wang2020minilm}, Jasper-0.6B \citep{zhang2025jasper}, Nemotron-1B and Qwen3-Embedding-8B \citep{yang2025qwen3}; Llama-3.1-8B \citep{grattafiori2024llama}; Skrub's TargetEncoder \citep{micci2001preprocessing}; and TARTE \citep{kim2025table}. The language models embed each cell on its own.
    \item \textbf{Tabular learners:} the table foundation models TabPFN-2.5 \citep{grinsztajn2025tabpfn}, TabPFN-3 \citep{grinsztajn2026tabpfn}, TabPFN-3.5 \citep{jager2026tabpfn}, TabICLv2 \citep{qu2026tabiclv2} and Causilo \citep{cho2026causilo}, and the learners trained from scratch TabM \citep{gorishniy2025tabm}, RealMLP \citep{holzmuller2024better}, XGBoost \citep{chen2016xgboost}, ExtraTrees and Ridge. The suffix ``-tuned'' marks a hyper-parameter-tuned learner.
    \item \textbf{End-to-end models}, which handle strings internally: CatBoost \citep{prokhorenkova2018catboost}, ContextTab \citep{spinaci2025contexttab}, TabSTAR \citep{arazi2025tabstar} and Mambular \citep{thielmann2024mambular}.
\end{itemize}

\begin{figure}[h]
    \centering
    \includegraphics[width=\linewidth]{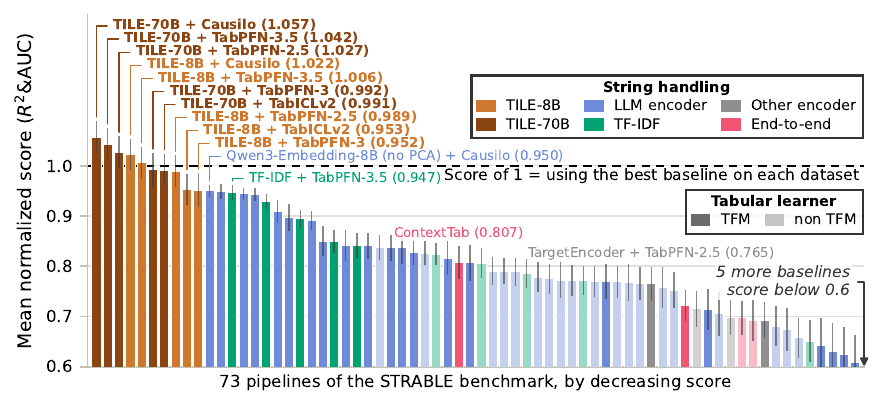}
    \caption{\textbf{A larger backbone further improves TILE on STRABLE.} Same as \autoref{fig:strable}, with TILE also run on Llama-3.1-70B (TILE-70B, dark brown); the baselines and their normalization are unchanged. With every learner, TILE with Llama-3.1-70B scores above TILE with Llama-3.1-8B.}
    \label{fig:strable_70b}
\end{figure}

\begin{table}[p]
    \caption{\textbf{Raw scores on STRABLE.} Unnormalized counterpart of \autoref{fig:strable}, multiplied by 100: the ROC-AUC averaged over the 32 classification datasets, and the $R^2$ (clipped at 0) averaged over the 76 regression datasets, each dataset scored by its average over 3 cross-validation folds. The $\pm$ is the standard error of that mean due to the cross-validation splits. TILE is run with Llama-3.1-8B and Llama-3.1-70B (\autoref{fig:strable_70b}). Labels in parentheses are STRABLE's preprocessing variants of the same pipeline. Best result in bold.}
    \label{tab:strable_raw}
    \centering
    \scriptsize
    \newcommand{\pmse}[1]{{\tiny$\pm$#1}}
    \setlength{\tabcolsep}{6pt}
    \renewcommand{\arraystretch}{0.9}
    \begin{tabular}{lcc}
        \toprule
        \textbf{Pipeline} & \textbf{Classification AUC (32)} & \textbf{Regression $R^2$ (76)} \\
        \midrule
        TILE (Llama-3.1-70B) + Causilo & 87.63 \pmse{0.11} & \textbf{74.23} \pmse{0.13} \\
        TILE (Llama-3.1-70B) + TabPFN-3.5 & \textbf{87.66} \pmse{0.10} & 73.63 \pmse{0.14} \\
        TILE (Llama-3.1-70B) + TabPFN-2.5 & 87.05 \pmse{0.11} & 73.65 \pmse{0.14} \\
        TILE (Llama-3.1-8B) + Causilo & 87.19 \pmse{0.12} & 73.41 \pmse{0.11} \\
        TILE (Llama-3.1-8B) + TabPFN-3.5 & 87.27 \pmse{0.11} & 72.67 \pmse{0.13} \\
        TILE (Llama-3.1-70B) + TabPFN-3 & 86.75 \pmse{0.12} & 72.67 \pmse{0.15} \\
        TILE (Llama-3.1-70B) + TabICLv2 & 87.05 \pmse{0.11} & 72.40 \pmse{0.15} \\
        TILE (Llama-3.1-8B) + TabPFN-2.5 & 86.59 \pmse{0.09} & 72.71 \pmse{0.13} \\
        TILE (Llama-3.1-8B) + TabICLv2 & 86.62 \pmse{0.11} & 71.41 \pmse{0.14} \\
        TILE (Llama-3.1-8B) + TabPFN-3 & 86.18 \pmse{0.12} & 71.80 \pmse{0.13} \\
        \midrule
        Qwen3-Embedding-8B (no PCA) + Causilo & 86.34 \pmse{0.11} & 71.52 \pmse{0.14} \\
        Qwen3-Embedding-8B (no PCA) + TabPFN-3.5 & 86.56 \pmse{0.11} & 71.41 \pmse{0.14} \\
        TF-IDF + TabPFN-3.5 & 86.75 \pmse{0.12} & 70.99 \pmse{0.16} \\
        Llama-3.1-8B (no PCA) + TabPFN-3.5 & 86.38 \pmse{0.11} & 71.25 \pmse{0.14} \\
        Llama-3.1-8B (no PCA) + Causilo & 86.18 \pmse{0.12} & 71.25 \pmse{0.14} \\
        TF-IDF + Causilo & 86.36 \pmse{0.13} & 70.66 \pmse{0.17} \\
        All-MiniLM-L6-v2 + TabPFN-3.5 & 86.33 \pmse{0.12} & 69.54 \pmse{0.14} \\
        All-MiniLM-L6-v2 + Causilo & 85.96 \pmse{0.13} & 69.38 \pmse{0.17} \\
        TF-IDF + TabPFN-2.5 & 85.93 \pmse{0.13} & 69.65 \pmse{0.16} \\
        Qwen3-Embedding-8B (no PCA) + TabPFN-2.5 & 85.50 \pmse{0.12} & 70.03 \pmse{0.14} \\
        E5-small-v2 + TabPFN-2.5 & 85.74 \pmse{0.12} & 67.46 \pmse{0.23} \\
        TF-IDF + TabPFN-3 & 85.19 \pmse{0.14} & 68.66 \pmse{0.15} \\
        FastText + TabPFN-2.5 & 85.24 \pmse{0.12} & 67.69 \pmse{0.19} \\
        TF-IDF + TabICLv2 & 85.38 \pmse{0.14} & 68.33 \pmse{0.17} \\
        Qwen3-Embedding-8B (no PCA) + TabPFN-3 & 84.75 \pmse{0.12} & 68.91 \pmse{0.14} \\
        Qwen3-Embedding-8B + TabM & 84.39 \pmse{0.11} & 68.36 \pmse{0.14} \\
        Qwen3-Embedding-8B (no PCA) + TabICLv2 & 85.06 \pmse{0.10} & 68.59 \pmse{0.15} \\
        All-MiniLM-L6-v2 + TabPFN-2.5 & 85.40 \pmse{0.15} & 67.22 \pmse{0.31} \\
        Llama-3.1-8B (no PCA) + TabPFN-3 & 84.43 \pmse{0.12} & 68.57 \pmse{0.14} \\
        Llama-3.1-8B + TabM & 84.10 \pmse{0.12} & 67.99 \pmse{0.15} \\
        TF-IDF + XGBoost-tuned & 85.01 \pmse{0.15} & 67.52 \pmse{0.15} \\
        Llama-3.1-8B (StandScal + PCA (30)) + TabPFN-2.5 & 85.83 \pmse{0.14} & 65.48 \pmse{0.28} \\
        ContextTab & 84.23 \pmse{0.10} & 68.11 \pmse{0.15} \\
        Nemotron-1B + TabPFN-2.5 & 85.37 \pmse{0.16} & 65.27 \pmse{0.29} \\
        TF-IDF + RealMLP & 83.81 \pmse{0.13} & 67.88 \pmse{0.18} \\
        All-MiniLM-L6-v2 + TabM & 83.57 \pmse{0.11} & 67.16 \pmse{0.14} \\
        E5-small-v2 + XGBoost-tuned & 84.71 \pmse{0.11} & 66.38 \pmse{0.16} \\
        E5-small-v2 + ExtraTrees-tuned & 84.51 \pmse{0.14} & 66.92 \pmse{0.16} \\
        TF-IDF + ExtraTrees-tuned & 84.54 \pmse{0.14} & 66.64 \pmse{0.18} \\
        All-MiniLM-L6-v2 + ExtraTrees-tuned & 84.28 \pmse{0.12} & 66.68 \pmse{0.17} \\
        FastText + XGBoost-tuned & 84.06 \pmse{0.20} & 66.25 \pmse{0.13} \\
        All-MiniLM-L6-v2 + RealMLP & 82.98 \pmse{0.09} & 67.27 \pmse{0.16} \\
        All-MiniLM-L6-v2 + XGBoost-tuned & 84.33 \pmse{0.14} & 65.98 \pmse{0.21} \\
        TF-IDF + TabM & 83.69 \pmse{0.11} & 66.09 \pmse{0.16} \\
        FastText + ExtraTrees-tuned & 83.82 \pmse{0.15} & 66.57 \pmse{0.19} \\
        All-MiniLM-L6-v2 + TabPFN-3 & 84.12 \pmse{0.12} & 66.21 \pmse{0.21} \\
        Llama-3.1-8B + ExtraTrees-tuned & 84.43 \pmse{0.13} & 66.21 \pmse{0.21} \\
        Llama-3.1-8B + RealMLP & 83.27 \pmse{0.15} & 66.32 \pmse{0.35} \\
        Qwen3-Embedding-8B + ExtraTrees-tuned & 84.65 \pmse{0.14} & 65.86 \pmse{0.21} \\
        TargetEncoder + TabPFN-2.5 & 82.84 \pmse{0.13} & 66.56 \pmse{0.16} \\
        Qwen3-Embedding-8B + RealMLP & 83.35 \pmse{0.13} & 66.61 \pmse{0.24} \\
        Jasper-0.6B + ExtraTrees-tuned & 84.49 \pmse{0.12} & 65.27 \pmse{0.20} \\
        TabSTAR & 83.74 \pmse{0.31} & 64.30 \pmse{0.22} \\
        TargetEncoder + XGBoost-tuned & 82.21 \pmse{0.16} & 65.24 \pmse{0.16} \\
        Qwen3-Embedding-8B + TabICLv2 & 84.71 \pmse{0.15} & 62.41 \pmse{0.33} \\
        E5-small-v2 + ExtraTrees & 81.22 \pmse{0.12} & 65.97 \pmse{0.17} \\
        TargetEncoder + ExtraTrees-tuned & 81.99 \pmse{0.15} & 64.85 \pmse{0.17} \\
        CatBoost & 82.18 \pmse{0.12} & 64.15 \pmse{0.17} \\
        CatBoost-tuned & 81.96 \pmse{0.17} & 64.22 \pmse{0.18} \\
        TARTE + TabPFN-2.5 & 81.33 \pmse{0.15} & 65.17 \pmse{0.15} \\
        Llama-3.1-8B + XGBoost-tuned & 84.37 \pmse{0.22} & 61.17 \pmse{0.53} \\
        Qwen3-Embedding-8B + XGBoost-tuned & 84.50 \pmse{0.13} & 61.12 \pmse{0.42} \\
        Jasper-0.6B + XGBoost-tuned & 83.92 \pmse{0.23} & 61.24 \pmse{0.40} \\
        TF-IDF + XGBoost & 82.66 \pmse{0.11} & 63.29 \pmse{0.16} \\
        Qwen3-Embedding-8B + TabPFN-2.5 & 85.07 \pmse{0.11} & 58.42 \pmse{0.46} \\
        Llama-3.1-8B + TabPFN-2.5 & 84.49 \pmse{0.14} & 57.84 \pmse{0.42} \\
        Jasper-0.6B + TabPFN-2.5 & 85.01 \pmse{0.12} & 57.92 \pmse{0.37} \\
        Llama-3.1-8B (StandScal + PCA (30)) + TabICLv2 & 84.29 \pmse{0.18} & 57.75 \pmse{0.51} \\
        Llama-3.1-8B + TabICLv2 & 84.38 \pmse{0.17} & 57.46 \pmse{0.55} \\
        TARTE + XGBoost-tuned & 79.77 \pmse{0.15} & 60.09 \pmse{0.13} \\
        Qwen3-Embedding-8B + Ridge & 80.79 \pmse{0.14} & 56.12 \pmse{0.15} \\
        TARTE + ExtraTrees-tuned & 78.32 \pmse{0.13} & 56.46 \pmse{0.13} \\
        Mambular & 76.32 \pmse{0.20} & 49.93 \pmse{0.19} \\
        \bottomrule
    \end{tabular}
\end{table}

Tables~\ref{tab:ttb_reg_1}--\ref{tab:wdbs_clf_3} report the raw $R^2$ and ROC-AUC scores of all methods on each individual dataset, averaged over 5 random seeds.

\begin{table}[h]
    \caption{\textbf{Regression results for TextTabBench datasets (part 1 of 3).} Raw $R^2$ scores (clipped at 0 if negative) multiplied by 100, averaged over 5 random seeds for a train size of 1,024 samples, for TabPFN-2.5, TabICLv2 and TabPFN-3, each fed with TILE, CASE-like, the cell-wise LLM encoder (Llama-3.1-8B) or TF-IDF. Best in bold, second best underlined, both computed over all 35 methods.}
    \label{tab:ttb_reg_1}
    \centering
    \scriptsize
    \setlength{\tabcolsep}{3.5pt}
    \resizebox{\linewidth}{!}{%
\begin{tabular}{l|cccc|cccc|cccc}
        \toprule
        \textbf{Method} & \multicolumn{4}{c}{\textbf{TabPFN-2.5}} & \multicolumn{4}{c}{\textbf{TabICLv2}} & \multicolumn{4}{c}{\textbf{TabPFN-3}} \\
        \textbf{String encoding} & TILE & CASE-like & LLM encoder & TF-IDF & TILE & CASE-like & LLM encoder & TF-IDF & TILE & CASE-like & LLM encoder & TF-IDF \\
        \midrule
        Airbnb & 73.98 & 72.38 & 55.48 & 73.77 & 71.00 & 69.86 & 52.16 & 71.66 & 71.70 & 70.80 & 64.30 & 71.24 \\
        Beer & 66.48 & 57.30 & 62.73 & 58.71 & 66.39 & 58.02 & 62.68 & 58.78 & 65.84 & 59.08 & 62.91 & 58.45 \\
        COVID Trials & 22.11 & 23.22 & 18.62 & 25.07 & 21.77 & 22.50 & 14.19 & 24.82 & 21.12 & 22.30 & 18.64 & 23.90 \\
        California Houses & 69.15 & 65.18 & 67.11 & 67.24 & 67.88 & 66.22 & 66.64 & 66.53 & 68.36 & 63.67 & 66.49 & 66.60 \\
        IT Salary & 47.66 & 46.77 & 39.18 & 46.93 & 46.64 & 44.35 & 0.84 & 33.50 & 47.72 & 38.81 & 42.39 & 44.74 \\
        Insurance Complaints & 42.75 & 39.68 & 35.48 & 42.34 & 44.58 & 44.46 & 35.87 & 44.30 & 44.30 & 43.69 & 35.85 & 42.98 \\
        Mercari & \underline{44.15} & 24.09 & 32.50 & 26.53 & 43.79 & 23.81 & 32.83 & 26.84 & 40.61 & 26.24 & 29.71 & 25.27 \\
        San Francisco Permits & 62.85 & 62.87 & 57.09 & 63.65 & 61.71 & 63.46 & 47.41 & 63.99 & 62.69 & 63.34 & 60.36 & 64.18 \\
        Stack Overflow & 99.73 & 99.82 & 85.96 & 99.67 & 99.73 & 99.78 & 87.65 & 99.86 & 99.62 & 97.08 & 79.10 & 99.70 \\
        Wine & 86.68 & 74.79 & 69.19 & 76.90 & 84.41 & 73.21 & 65.46 & 73.73 & 85.58 & 75.50 & 71.08 & 74.92 \\
        \bottomrule
    \end{tabular}}
\end{table}

\begin{table}[h]
    \caption{\textbf{Regression results for TextTabBench datasets (part 2 of 3).} Raw $R^2$ scores (clipped at 0 if negative) multiplied by 100, averaged over 5 random seeds for a train size of 1,024 samples, for TabPFN-3.5, EXAONE-Tabular and Causilo, each fed with TILE, CASE-like, the cell-wise LLM encoder (Llama-3.1-8B) or TF-IDF. Best in bold, second best underlined, both computed over all 35 methods.}
    \label{tab:ttb_reg_2}
    \centering
    \scriptsize
    \setlength{\tabcolsep}{3.5pt}
    \resizebox{\linewidth}{!}{%
    \begin{tabular}{l|cccc|cccc|cccc}
        \toprule
        \textbf{Method} & \multicolumn{4}{c}{\textbf{TabPFN-3.5}} & \multicolumn{4}{c}{\textbf{EXAONE-Tabular}} & \multicolumn{4}{c}{\textbf{Causilo}} \\
        \textbf{String encoding} & TILE & CASE-like & LLM encoder & TF-IDF & TILE & CASE-like & LLM encoder & TF-IDF & TILE & CASE-like & LLM encoder & TF-IDF \\
        \midrule
        Airbnb & 72.29 & 70.12 & 66.51 & 73.15 & \textbf{74.36} & 73.57 & 64.80 & \underline{74.27} & 74.15 & 73.30 & 64.22 & 74.21 \\
        Beer & \underline{66.71} & 64.28 & 64.45 & 61.84 & \textbf{67.28} & 59.20 & 64.30 & 60.66 & 66.67 & 59.75 & 64.02 & 59.32 \\
        COVID Trials & 24.73 & 26.13 & 23.90 & \underline{26.32} & 22.88 & 24.81 & 22.51 & 25.58 & 25.91 & 26.21 & 20.01 & \textbf{26.76} \\
        California Houses & 69.22 & 66.28 & 66.92 & 68.01 & \textbf{70.66} & 68.51 & 68.13 & 68.51 & \underline{69.88} & 68.07 & 68.10 & 68.14 \\
        IT Salary & \underline{48.64} & 46.43 & 38.28 & 45.50 & 47.32 & 46.54 & 41.60 & 45.44 & \textbf{48.82} & 44.38 & 20.59 & 29.52 \\
        Insurance Complaints & 44.62 & 45.79 & 26.50 & \underline{45.98} & 44.61 & 44.46 & 40.51 & 45.39 & 44.30 & 45.46 & 39.28 & 44.65 \\
        Mercari & 43.51 & 31.58 & 34.47 & 29.91 & 43.36 & 26.67 & 35.13 & 29.58 & \textbf{45.23} & 29.74 & 34.63 & 29.36 \\
        San Francisco Permits & 62.47 & 63.67 & 57.83 & \underline{64.37} & 62.89 & 63.68 & 60.91 & 64.20 & 63.40 & 63.91 & 61.15 & \textbf{64.47} \\
        Stack Overflow & 99.79 & 99.81 & 77.97 & 99.78 & 99.93 & \textbf{99.94} & 99.88 & \underline{99.93} & 99.87 & 99.89 & 99.26 & 99.86 \\
        Wine & 86.51 & 77.83 & 78.81 & 77.63 & \underline{87.17} & 76.19 & 76.96 & 78.33 & \textbf{87.42} & 74.86 & 79.91 & 77.29 \\
        \bottomrule
    \end{tabular}}
\end{table}

\begin{table}[h]
    \caption{\textbf{Regression results for TextTabBench datasets (part 3 of 3).} Raw $R^2$ scores (clipped at 0 if negative) multiplied by 100, averaged over 5 random seeds for a train size of 1,024 samples, for the remaining learners, fed with the cell-wise LLM encoder (Llama-3.1-8B) or TF-IDF, and the methods with internal string handling. Best in bold, second best underlined, both computed over all 35 methods.}
    \label{tab:ttb_reg_3}
    \centering
    \scriptsize
    \setlength{\tabcolsep}{3.5pt}
    \resizebox{\linewidth}{!}{%
    \begin{tabular}{l|cc|ccc|cc|cc|c|c}
        \toprule
        \textbf{Method} & \multicolumn{2}{c}{\textbf{XGBoost (tuned)}} & \multicolumn{3}{c}{\textbf{CatBoost}} & \multicolumn{2}{c}{\textbf{RealMLP (tuned)}} & \multicolumn{2}{c}{\textbf{Ridge}} & \textbf{ContextTab} & \textbf{TabSTAR} \\
        \textbf{String encoding} & LLM encoder & TF-IDF & LLM encoder & TF-IDF & internal & LLM encoder & TF-IDF & LLM encoder & TF-IDF & internal & internal \\
        \midrule
        Airbnb & 60.09 & 71.36 & 48.67 & 72.27 & 71.90 & 65.84 & 71.47 & 62.41 & 68.56 & 72.74 & 69.71 \\
        Beer & 59.74 & 56.62 & 60.18 & 56.38 & 56.55 & 61.33 & 56.68 & 53.68 & 48.06 & 62.22 & 49.16 \\
        COVID Trials & 15.25 & 21.50 & 13.18 & 22.67 & 18.39 & 13.15 & 16.48 & 4.07 & 14.99 & 15.77 & 14.98 \\
        California Houses & 61.95 & 62.50 & 59.23 & 60.85 & 62.03 & 62.03 & 62.00 & 53.55 & 58.26 & 65.82 & 64.72 \\
        IT Salary & 28.56 & 36.56 & 24.09 & 33.31 & 43.51 & 33.82 & 37.86 & 25.08 & 28.13 & 45.77 & 41.59 \\
        Insurance Complaints & 32.40 & 41.74 & 29.68 & 42.71 & 42.79 & 28.58 & 40.83 & 36.90 & 37.19 & \textbf{48.72} & 32.26 \\
        Mercari & 30.62 & 25.55 & 30.63 & 25.43 & 15.53 & 31.38 & 24.86 & 25.72 & 21.42 & 26.93 & 41.42 \\
        San Francisco Permits & 53.23 & 62.59 & 51.92 & 61.86 & 59.84 & 36.24 & 62.40 & 54.77 & 54.11 & 60.27 & 59.41 \\
        Stack Overflow & 96.85 & 97.90 & 94.04 & 98.22 & 89.78 & 90.71 & 97.47 & 31.92 & 41.50 & 82.93 & 97.53 \\
        Wine & 71.74 & 70.34 & 65.02 & 68.92 & 69.20 & 68.77 & 74.97 & 76.15 & 69.35 & 73.04 & 67.75 \\
        \bottomrule
    \end{tabular}}
\end{table}

\FloatBarrier

\begin{table}[h]
    \caption{\textbf{Regression results for CARTE datasets (part 1 of 3).} Raw $R^2$ scores (clipped at 0 if negative) multiplied by 100, averaged over 5 random seeds for a train size of 1,024 samples, for TabPFN-2.5, TabICLv2 and TabPFN-3, each fed with TILE, CASE-like, the cell-wise LLM encoder (Llama-3.1-8B) or TF-IDF. Best in bold, second best underlined, both computed over all 35 methods.}
    \label{tab:carte_reg_1}
    \centering
    \scriptsize
    \setlength{\tabcolsep}{3.5pt}
    \resizebox{\linewidth}{!}{%
    \begin{tabular}{l|cccc|cccc|cccc}
        \toprule
        \textbf{Method} & \multicolumn{4}{c}{\textbf{TabPFN-2.5}} & \multicolumn{4}{c}{\textbf{TabICLv2}} & \multicolumn{4}{c}{\textbf{TabPFN-3}} \\
        \textbf{String encoding} & TILE & CASE-like & LLM encoder & TF-IDF & TILE & CASE-like & LLM encoder & TF-IDF & TILE & CASE-like & LLM encoder & TF-IDF \\
        \midrule
        Anime Planet & 64.58 & 46.31 & 54.11 & 54.43 & 63.28 & 39.48 & 56.66 & 54.29 & 63.54 & 45.92 & 57.03 & 54.52 \\
        Babies R Us & 81.85 & 76.26 & 53.45 & 79.38 & 82.70 & 71.75 & 68.73 & 79.39 & 82.47 & 77.76 & 70.49 & 79.84 \\
        Beer Ratings & 60.94 & 51.78 & 57.67 & 53.38 & 60.57 & 52.03 & 57.78 & 54.36 & 59.66 & 52.80 & 56.61 & 52.32 \\
        BikeDekho & \textbf{81.10} & 78.12 & 63.47 & 78.81 & 79.49 & 77.75 & 71.36 & 78.66 & \underline{80.08} & 78.85 & 74.62 & 78.73 \\
        BikeWale & \textbf{81.12} & 72.46 & 76.74 & 75.04 & 80.30 & 72.70 & 77.14 & 75.32 & 80.50 & 75.36 & 78.82 & 74.96 \\
        Buy Buy Baby & 82.06 & 76.43 & 68.66 & 77.79 & 81.86 & 74.99 & 77.98 & 78.22 & 81.80 & 78.71 & 78.86 & 77.76 \\
\bottomrule
    \end{tabular}}
\end{table}

\begin{table}[h]
    \caption{\textbf{Regression results for CARTE datasets (part 2 of 3).} Raw $R^2$ scores (clipped at 0 if negative) multiplied by 100, averaged over 5 random seeds for a train size of 1,024 samples, for TabPFN-3.5, EXAONE-Tabular and Causilo, each fed with TILE, CASE-like, the cell-wise LLM encoder (Llama-3.1-8B) or TF-IDF. Best in bold, second best underlined, both computed over all 35 methods.}
    \label{tab:carte_reg_2}
    \centering
    \scriptsize
    \setlength{\tabcolsep}{3.5pt}
    \resizebox{\linewidth}{!}{%
    \begin{tabular}{l|cccc|cccc|cccc}
        \toprule
        \textbf{Method} & \multicolumn{4}{c}{\textbf{TabPFN-3.5}} & \multicolumn{4}{c}{\textbf{EXAONE-Tabular}} & \multicolumn{4}{c}{\textbf{Causilo}} \\
        \textbf{String encoding} & TILE & CASE-like & LLM encoder & TF-IDF & TILE & CASE-like & LLM encoder & TF-IDF & TILE & CASE-like & LLM encoder & TF-IDF \\
        \midrule
        Anime Planet & 64.43 & 50.64 & 59.31 & 55.22 & \textbf{65.49} & 48.39 & 58.93 & 55.93 & \underline{65.19} & 48.04 & 59.41 & 55.42 \\
        Babies R Us & 82.87 & 80.86 & 73.11 & 81.67 & 82.92 & 78.95 & 73.90 & 79.73 & \underline{83.69} & 80.35 & 70.64 & 81.19 \\
        Beer Ratings & 60.32 & 56.35 & 58.86 & 56.52 & \textbf{61.22} & 53.08 & 59.23 & 55.36 & \underline{61.21} & 53.37 & 58.87 & 54.57 \\
        BikeDekho & 79.49 & 79.95 & 74.81 & 78.85 & 79.61 & 78.16 & 73.78 & 77.90 & 79.56 & 79.28 & 70.28 & 79.11 \\
        BikeWale & 80.58 & 76.53 & 79.00 & 76.44 & \underline{80.89} & 78.82 & 78.25 & 75.90 & 80.77 & 79.61 & 77.80 & 75.73 \\
        Buy Buy Baby & \textbf{82.59} & 81.50 & 80.37 & 78.66 & \underline{82.38} & 79.25 & 78.25 & 79.13 & 82.00 & 80.08 & 78.03 & 79.26 \\
\bottomrule
    \end{tabular}}
\end{table}

\begin{table}[h]
    \caption{\textbf{Regression results for CARTE datasets (part 3 of 3).} Raw $R^2$ scores (clipped at 0 if negative) multiplied by 100, averaged over 5 random seeds for a train size of 1,024 samples, for the remaining learners, fed with the cell-wise LLM encoder (Llama-3.1-8B) or TF-IDF, and the methods with internal string handling. Best in bold, second best underlined, both computed over all 35 methods.}
    \label{tab:carte_reg_3}
    \centering
    \scriptsize
    \setlength{\tabcolsep}{3.5pt}
    \resizebox{\linewidth}{!}{%
    \begin{tabular}{l|cc|ccc|cc|cc|c|c}
        \toprule
        \textbf{Method} & \multicolumn{2}{c}{\textbf{XGBoost (tuned)}} & \multicolumn{3}{c}{\textbf{CatBoost}} & \multicolumn{2}{c}{\textbf{RealMLP (tuned)}} & \multicolumn{2}{c}{\textbf{Ridge}} & \textbf{ContextTab} & \textbf{TabSTAR} \\
        \textbf{String encoding} & LLM encoder & TF-IDF & LLM encoder & TF-IDF & internal & LLM encoder & TF-IDF & LLM encoder & TF-IDF & internal & internal \\
        \midrule
        Anime Planet & 55.91 & 50.86 & 50.86 & 50.23 & 45.05 & 54.56 & 50.39 & 56.03 & 47.90 & 53.09 & 55.79 \\
        Babies R Us & 55.41 & 74.83 & 51.23 & 74.65 & 54.22 & 58.82 & 77.73 & 58.43 & 62.45 & 74.28 & \textbf{85.33} \\
        Beer Ratings & 53.63 & 50.72 & 53.82 & 50.51 & 50.76 & 54.48 & 49.88 & 49.15 & 40.39 & 56.77 & 36.91 \\
        BikeDekho & 66.46 & 75.47 & 64.53 & 76.17 & 65.78 & 68.16 & 75.24 & 69.76 & 38.79 & 77.25 & 76.17 \\
        BikeWale & 73.11 & 70.53 & 69.96 & 70.60 & 55.43 & 70.22 & 72.23 & 70.15 & 47.20 & 71.47 & 78.68 \\
        Buy Buy Baby & 77.45 & 76.98 & 67.67 & 76.47 & 66.89 & 77.52 & 75.96 & 68.41 & 62.21 & 78.38 & 73.73 \\
\bottomrule
    \end{tabular}}
\end{table}

\FloatBarrier

\begin{table}[h]
    \caption{\textbf{Regression results for WikiDBs datasets (part 1 of 3).} Raw $R^2$ scores (clipped at 0 if negative) multiplied by 100, averaged over 5 random seeds for a train size of 1,024 samples, for TabPFN-2.5, TabICLv2 and TabPFN-3, each fed with TILE, CASE-like, the cell-wise LLM encoder (Llama-3.1-8B) or TF-IDF. Best in bold, second best underlined, both computed over all 35 methods.}
    \label{tab:wdbs_reg_1}
    \centering
    \scriptsize
    \setlength{\tabcolsep}{3.5pt}
    \resizebox{\linewidth}{!}{%
    \begin{tabular}{l|cccc|cccc|cccc}
        \toprule
        \textbf{Method} & \multicolumn{4}{c}{\textbf{TabPFN-2.5}} & \multicolumn{4}{c}{\textbf{TabICLv2}} & \multicolumn{4}{c}{\textbf{TabPFN-3}} \\
        \textbf{String encoding} & TILE & CASE-like & LLM encoder & TF-IDF & TILE & CASE-like & LLM encoder & TF-IDF & TILE & CASE-like & LLM encoder & TF-IDF \\
        \midrule
        Artworks Inventory & 32.21 & 27.13 & 25.01 & 28.71 & 32.79 & 28.18 & 26.49 & 29.11 & 32.64 & 27.76 & 28.13 & 30.01 \\
        Business Locations & \underline{87.73} & 77.36 & 78.46 & 72.02 & 84.64 & 66.91 & 71.78 & 61.73 & 86.62 & 78.16 & 77.07 & 70.58 \\
        Dissolved Municipalities & 64.16 & 37.54 & 56.63 & 54.56 & 64.49 & 46.09 & 56.10 & 53.42 & 63.70 & 41.34 & 56.60 & 53.92 \\
        Drawings Catalog & 80.63 & 80.23 & 80.49 & 80.58 & 80.73 & 79.93 & 80.44 & 80.24 & 80.80 & 80.60 & 79.95 & 80.53 \\
        Eclipsing Binary Stars & 63.99 & 27.11 & 62.06 & 60.07 & 63.00 & 0.00 & 60.14 & 58.61 & 63.44 & 13.68 & 61.97 & 59.87 \\
        Geopolitical Regions & \textbf{66.89} & 53.63 & 55.40 & 50.08 & 64.51 & 28.50 & 49.97 & 45.14 & 62.67 & 52.32 & 51.89 & 45.54 \\
\bottomrule
    \end{tabular}}
\end{table}

\begin{table}[h]
    \caption{\textbf{Regression results for WikiDBs datasets (part 2 of 3).} Raw $R^2$ scores (clipped at 0 if negative) multiplied by 100, averaged over 5 random seeds for a train size of 1,024 samples, for TabPFN-3.5, EXAONE-Tabular and Causilo, each fed with TILE, CASE-like, the cell-wise LLM encoder (Llama-3.1-8B) or TF-IDF. Best in bold, second best underlined, both computed over all 35 methods.}
    \label{tab:wdbs_reg_2}
    \centering
    \scriptsize
    \setlength{\tabcolsep}{3.5pt}
    \resizebox{\linewidth}{!}{%
    \begin{tabular}{l|cccc|cccc|cccc}
        \toprule
        \textbf{Method} & \multicolumn{4}{c}{\textbf{TabPFN-3.5}} & \multicolumn{4}{c}{\textbf{EXAONE-Tabular}} & \multicolumn{4}{c}{\textbf{Causilo}} \\
        \textbf{String encoding} & TILE & CASE-like & LLM encoder & TF-IDF & TILE & CASE-like & LLM encoder & TF-IDF & TILE & CASE-like & LLM encoder & TF-IDF \\
        \midrule
        Artworks Inventory & 32.96 & 29.08 & 26.60 & 29.91 & \underline{33.01} & 29.10 & 31.10 & 30.19 & \textbf{33.28} & 30.11 & 29.95 & 30.17 \\
        Business Locations & \textbf{87.93} & 79.49 & 78.79 & 72.92 & 86.97 & 74.44 & 78.30 & 71.16 & 87.24 & 70.23 & 77.58 & 69.15 \\
        Dissolved Municipalities & \textbf{65.74} & 46.76 & 58.34 & 55.88 & 64.97 & 54.18 & 58.66 & 55.69 & \underline{65.50} & 53.60 & 57.85 & 55.37 \\
        Drawings Catalog & 81.30 & \textbf{81.68} & 81.26 & 81.23 & 81.20 & 81.22 & 80.85 & \underline{81.50} & 81.37 & 81.30 & 80.84 & 81.02 \\
        Eclipsing Binary Stars & 63.56 & 57.36 & 61.82 & 59.37 & \underline{64.15} & 61.45 & 61.96 & 60.47 & \textbf{64.21} & 60.42 & 62.34 & 60.62 \\
        Geopolitical Regions & 64.39 & 56.56 & 56.39 & 50.28 & \underline{66.64} & 36.73 & 56.05 & 53.43 & 64.90 & 34.45 & 56.42 & 51.06 \\
\bottomrule
    \end{tabular}}
\end{table}

\begin{table}[h]
    \caption{\textbf{Regression results for WikiDBs datasets (part 3 of 3).} Raw $R^2$ scores (clipped at 0 if negative) multiplied by 100, averaged over 5 random seeds for a train size of 1,024 samples, for the remaining learners, fed with the cell-wise LLM encoder (Llama-3.1-8B) or TF-IDF, and the methods with internal string handling. Best in bold, second best underlined, both computed over all 35 methods.}
    \label{tab:wdbs_reg_3}
    \centering
    \scriptsize
    \setlength{\tabcolsep}{3.5pt}
    \resizebox{\linewidth}{!}{%
    \begin{tabular}{l|cc|ccc|cc|cc|c|c}
        \toprule
        \textbf{Method} & \multicolumn{2}{c}{\textbf{XGBoost (tuned)}} & \multicolumn{3}{c}{\textbf{CatBoost}} & \multicolumn{2}{c}{\textbf{RealMLP (tuned)}} & \multicolumn{2}{c}{\textbf{Ridge}} & \textbf{ContextTab} & \textbf{TabSTAR} \\
        \textbf{String encoding} & LLM encoder & TF-IDF & LLM encoder & TF-IDF & internal & LLM encoder & TF-IDF & LLM encoder & TF-IDF & internal & internal \\
        \midrule
        Artworks Inventory & 28.24 & 26.32 & 15.49 & 21.32 & 24.82 & 18.26 & 19.18 & 19.36 & 10.86 & 28.66 & 28.23 \\
        Business Locations & 74.57 & 70.45 & 73.63 & 69.25 & 69.78 & 75.22 & 65.22 & 72.31 & 51.89 & 72.30 & 66.10 \\
        Dissolved Municipalities & 50.72 & 52.56 & 52.25 & 53.27 & 46.94 & 53.50 & 49.12 & 45.22 & 43.26 & 56.92 & 50.97 \\
        Drawings Catalog & 78.93 & 78.89 & 67.25 & 78.05 & 78.29 & 73.80 & 76.87 & 69.62 & 65.16 & 78.87 & 75.74 \\
        Eclipsing Binary Stars & 59.42 & 58.34 & 57.96 & 57.42 & 48.29 & 57.29 & 57.43 & 60.11 & 50.47 & 60.81 & 55.52 \\
        Geopolitical Regions & 49.99 & 46.68 & 50.20 & 48.47 & 48.02 & 46.47 & 43.14 & 41.04 & 23.83 & 50.60 & 49.05 \\
\bottomrule
    \end{tabular}}
\end{table}

\FloatBarrier

\begin{table}[h]
    \caption{\textbf{Classification results for TextTabBench datasets (part 1 of 3).} Raw ROC AUC scores multiplied by 100, averaged over 5 random seeds for a train size of 1,024 samples, for TabPFN-2.5, TabICLv2 and TabPFN-3, each fed with TILE, CASE-like, the cell-wise LLM encoder (Llama-3.1-8B) or TF-IDF. Best in bold, second best underlined, both computed over all 36 methods.}
    \label{tab:ttb_clf_1}
    \centering
    \scriptsize
    \setlength{\tabcolsep}{3.5pt}
    \resizebox{\linewidth}{!}{%
    \begin{tabular}{l|cccc|cccc|cccc}
        \toprule
        \textbf{Method} & \multicolumn{4}{c}{\textbf{TabPFN-2.5}} & \multicolumn{4}{c}{\textbf{TabICLv2}} & \multicolumn{4}{c}{\textbf{TabPFN-3}} \\
        \textbf{String encoding} & TILE & CASE-like & LLM encoder & TF-IDF & TILE & CASE-like & LLM encoder & TF-IDF & TILE & CASE-like & LLM encoder & TF-IDF \\
        \midrule
        Customer Complaints & 89.10 & 88.61 & 89.23 & 89.41 & 89.13 & 88.57 & 88.90 & 89.19 & 89.01 & 88.60 & 88.33 & 88.89 \\
        Diabetes & 97.62 & 97.70 & 97.69 & 97.72 & 97.64 & 97.63 & 97.70 & 97.77 & 97.83 & 97.79 & 97.89 & 97.87 \\
        Job Frauds & 99.15 & 97.82 & 98.69 & 98.73 & \underline{99.25} & 97.32 & 98.99 & 98.76 & 99.13 & 97.43 & 98.59 & 98.65 \\
        Kickstarter & 86.74 & 78.65 & 85.05 & 84.67 & 86.91 & 85.33 & 85.18 & 84.91 & 87.05 & 79.64 & 84.62 & 84.29 \\
        Lending Club & 68.96 & 69.48 & 69.20 & 69.16 & 69.91 & 69.64 & 69.01 & 69.26 & \underline{70.26} & 69.48 & 68.98 & 68.52 \\
        OSHA Accidents & 58.76 & 58.84 & 58.54 & 59.28 & 60.30 & 57.77 & 59.12 & 59.32 & 60.02 & 58.77 & 58.04 & 59.25 \\
        Spotify & 98.36 & 94.87 & 97.83 & 95.92 & 98.28 & 95.34 & 97.66 & 95.96 & 98.40 & 94.79 & 97.54 & 95.55 \\
        \bottomrule
    \end{tabular}}
\end{table}

\begin{table}[h]
    \caption{\textbf{Classification results for TextTabBench datasets (part 2 of 3).} Raw ROC AUC scores multiplied by 100, averaged over 5 random seeds for a train size of 1,024 samples, for TabPFN-3.5, EXAONE-Tabular and Causilo, each fed with TILE, CASE-like, the cell-wise LLM encoder (Llama-3.1-8B) or TF-IDF. Best in bold, second best underlined, both computed over all 36 methods.}
    \label{tab:ttb_clf_2}
    \centering
    \scriptsize
    \setlength{\tabcolsep}{3.5pt}
    \resizebox{\linewidth}{!}{%
    \begin{tabular}{l|cccc|cccc|cccc}
        \toprule
        \textbf{Method} & \multicolumn{4}{c}{\textbf{TabPFN-3.5}} & \multicolumn{4}{c}{\textbf{EXAONE-Tabular}} & \multicolumn{4}{c}{\textbf{Causilo}} \\
        \textbf{String encoding} & TILE & CASE-like & LLM encoder & TF-IDF & TILE & CASE-like & LLM encoder & TF-IDF & TILE & CASE-like & LLM encoder & TF-IDF \\
        \midrule
        Customer Complaints & 89.23 & 89.06 & 88.80 & 89.48 & 89.11 & 89.17 & \underline{89.65} & \textbf{89.86} & 89.32 & 88.92 & 89.08 & 89.45 \\
        Diabetes & 97.90 & 97.90 & 97.89 & \textbf{98.01} & 97.83 & 97.82 & 97.87 & \underline{97.92} & 97.89 & 97.85 & 97.87 & 97.90 \\
        Job Frauds & 99.21 & 99.04 & 98.85 & 98.92 & 99.12 & 98.78 & 99.04 & 98.79 & \textbf{99.27} & 98.71 & 98.70 & 98.81 \\
        Kickstarter & \textbf{87.36} & 86.07 & 86.15 & 85.93 & 86.47 & 86.03 & 86.23 & 85.48 & \underline{87.36} & 85.33 & 85.62 & 85.03 \\
        Lending Club & \textbf{70.37} & 69.68 & 69.96 & 69.31 & 68.73 & 69.71 & 69.43 & 69.43 & 69.48 & 69.74 & 68.99 & 69.27 \\
        OSHA Accidents & \underline{60.77} & 59.19 & 58.16 & 59.85 & 59.15 & 59.85 & 58.66 & 58.84 & \textbf{60.85} & 60.06 & 58.55 & 59.32 \\
        Spotify & \underline{98.62} & 97.40 & 98.15 & 96.99 & 98.47 & 96.17 & 98.00 & 96.90 & \textbf{98.62} & 96.49 & 98.06 & 96.26 \\
        \bottomrule
    \end{tabular}}
\end{table}

\begin{table}[h]
    \caption{\textbf{Classification results for TextTabBench datasets (part 3 of 3).} Raw ROC AUC scores multiplied by 100, averaged over 5 random seeds for a train size of 1,024 samples, for the remaining learners, fed with the cell-wise LLM encoder (Llama-3.1-8B) or TF-IDF, and the methods with internal string handling; TabLLM has no result on two datasets (--). Best in bold, second best underlined, both computed over all 36 methods.}
    \label{tab:ttb_clf_3}
    \centering
    \scriptsize
    \setlength{\tabcolsep}{3.5pt}
    \resizebox{\linewidth}{!}{%
    \begin{tabular}{l|cc|ccc|cc|cc|c|c|c}
        \toprule
        \textbf{Method} & \multicolumn{2}{c}{\textbf{XGBoost (tuned)}} & \multicolumn{3}{c}{\textbf{CatBoost}} & \multicolumn{2}{c}{\textbf{RealMLP (tuned)}} & \multicolumn{2}{c}{\textbf{Ridge}} & \textbf{ContextTab} & \textbf{TabSTAR} & \textbf{TabLLM} \\
        \textbf{String encoding} & LLM encoder & TF-IDF & LLM encoder & TF-IDF & internal & LLM encoder & TF-IDF & LLM encoder & TF-IDF & internal & internal & internal \\
        \midrule
        Customer Complaints & 89.16 & 89.37 & 87.83 & 88.08 & 88.43 & 86.99 & 88.00 & 81.59 & 85.20 & 88.46 & 87.67 & 87.40 \\
        Diabetes & 97.27 & 97.30 & 97.14 & 97.21 & 97.67 & 96.34 & 96.71 & 95.57 & 96.12 & 97.87 & 97.48 & 97.46 \\
        Job Frauds & 97.63 & 97.56 & 98.21 & 98.10 & 98.27 & 98.51 & 98.18 & 96.49 & 96.92 & 98.92 & 97.13 & -- \\
        Kickstarter & 82.90 & 82.00 & 82.44 & 81.76 & 84.62 & 80.41 & 79.63 & 76.32 & 73.65 & 83.38 & 73.24 & 80.49 \\
        Lending Club & 68.47 & 67.74 & 65.81 & 65.91 & 68.63 & 66.43 & 66.33 & 65.49 & 69.31 & 68.38 & 66.74 & -- \\
        OSHA Accidents & 57.67 & 57.97 & 54.52 & 57.37 & 56.84 & 54.69 & 57.82 & 58.06 & 58.35 & 57.99 & 57.94 & 58.76 \\
        Spotify & 96.26 & 93.51 & 96.52 & 94.28 & 93.88 & 95.81 & 92.94 & 91.81 & 88.22 & 96.44 & 92.94 & 96.86 \\
        \bottomrule
    \end{tabular}}
\end{table}

\FloatBarrier

\begin{table}[h]
    \caption{\textbf{Classification results for CARTE datasets (part 1 of 3).} Raw ROC AUC scores multiplied by 100, averaged over 5 random seeds for a train size of 1,024 samples, for TabPFN-2.5, TabICLv2 and TabPFN-3, each fed with TILE, CASE-like, the cell-wise LLM encoder (Llama-3.1-8B) or TF-IDF. Best in bold, second best underlined, both computed over all 36 methods.}
    \label{tab:carte_clf_1}
    \centering
    \scriptsize
    \setlength{\tabcolsep}{3.5pt}
    \resizebox{\linewidth}{!}{%
    \begin{tabular}{l|cccc|cccc|cccc}
        \toprule
        \textbf{Method} & \multicolumn{4}{c}{\textbf{TabPFN-2.5}} & \multicolumn{4}{c}{\textbf{TabICLv2}} & \multicolumn{4}{c}{\textbf{TabPFN-3}} \\
        \textbf{String encoding} & TILE & CASE-like & LLM encoder & TF-IDF & TILE & CASE-like & LLM encoder & TF-IDF & TILE & CASE-like & LLM encoder & TF-IDF \\
        \midrule
        Chocolate Bar Ratings & 81.87 & 78.28 & 80.17 & 80.62 & 81.98 & 78.80 & 80.71 & 80.29 & 81.06 & 78.50 & 78.47 & 79.06 \\
        Coffee Ratings & 92.90 & 84.68 & 91.81 & 93.48 & 92.44 & 85.27 & 92.45 & 93.44 & 92.08 & 84.67 & 90.85 & 92.51 \\
        Michelin & 90.66 & 84.69 & 89.07 & 88.20 & \underline{90.81} & 86.42 & 89.20 & 88.01 & 90.65 & 86.05 & 88.45 & 87.70 \\
        NBA Draft & 90.78 & 89.38 & 89.25 & 89.05 & 90.97 & 89.32 & 88.99 & 88.98 & 90.97 & 89.37 & 88.90 & 88.89 \\
        Ramen Ratings & 69.61 & 68.15 & 69.57 & 69.05 & 70.55 & 69.45 & 70.33 & 69.15 & 70.21 & 69.03 & 68.26 & 67.76 \\
        Roger Ebert & 83.76 & 78.53 & 78.82 & 75.84 & 84.00 & 78.31 & 78.44 & 75.69 & 84.05 & 78.36 & 78.21 & 75.74 \\
        Spotify & \textbf{98.93} & 94.94 & 95.42 & 90.57 & 98.72 & 94.95 & 95.55 & 90.23 & 98.68 & 94.86 & 94.76 & 90.37 \\
        US Accidents Severity & 79.44 & 77.64 & 78.03 & 78.51 & 78.46 & 78.16 & 77.92 & 78.06 & 79.26 & 77.76 & 75.12 & 77.31 \\
        Whisky & 88.04 & 86.52 & 86.99 & 86.50 & \textbf{88.32} & 86.72 & 87.20 & 86.63 & 88.05 & 86.70 & 85.22 & 86.07 \\
        Yelp & 80.58 & 79.54 & 78.10 & 76.45 & 80.62 & 79.36 & 77.91 & 76.00 & \underline{80.81} & 79.29 & 76.52 & 76.08 \\
        Zomato & 72.78 & 70.94 & 71.40 & 71.39 & 73.03 & 71.27 & 71.04 & 70.90 & 72.73 & 70.82 & 70.56 & 70.31 \\
        \bottomrule
    \end{tabular}}
\end{table}

\begin{table}[h]
    \caption{\textbf{Classification results for CARTE datasets (part 2 of 3).} Raw ROC AUC scores multiplied by 100, averaged over 5 random seeds for a train size of 1,024 samples, for TabPFN-3.5, EXAONE-Tabular and Causilo, each fed with TILE, CASE-like, the cell-wise LLM encoder (Llama-3.1-8B) or TF-IDF. Best in bold, second best underlined, both computed over all 36 methods.}
    \label{tab:carte_clf_2}
    \centering
    \scriptsize
    \setlength{\tabcolsep}{3.5pt}
    \resizebox{\linewidth}{!}{%
    \begin{tabular}{l|cccc|cccc|cccc}
        \toprule
        \textbf{Method} & \multicolumn{4}{c}{\textbf{TabPFN-3.5}} & \multicolumn{4}{c}{\textbf{EXAONE-Tabular}} & \multicolumn{4}{c}{\textbf{Causilo}} \\
        \textbf{String encoding} & TILE & CASE-like & LLM encoder & TF-IDF & TILE & CASE-like & LLM encoder & TF-IDF & TILE & CASE-like & LLM encoder & TF-IDF \\
        \midrule
        Chocolate Bar Ratings & \underline{82.39} & 80.17 & 80.98 & 81.58 & 82.33 & 78.44 & 81.06 & 81.26 & \textbf{82.43} & 79.06 & 80.67 & 80.60 \\
        Coffee Ratings & 92.81 & 87.04 & 92.57 & 93.58 & 92.94 & 85.97 & 91.91 & 93.73 & 93.12 & 85.51 & 92.39 & 93.37 \\
        Michelin & 90.77 & 87.19 & 88.66 & 88.31 & 90.52 & 87.32 & 89.45 & 88.23 & \textbf{90.84} & 87.19 & 88.94 & 88.12 \\
        NBA Draft & \underline{91.13} & 89.25 & 89.21 & 88.99 & 90.68 & 89.08 & 89.17 & 89.06 & \textbf{91.20} & 89.37 & 89.32 & 89.06 \\
        Ramen Ratings & 70.17 & 70.05 & 70.34 & 70.13 & 70.03 & 69.28 & 70.72 & 69.21 & \underline{70.72} & 69.55 & 70.46 & 69.31 \\
        Roger Ebert & 84.04 & 78.72 & 78.99 & 76.04 & \underline{84.16} & 77.74 & 78.77 & 75.90 & \textbf{84.28} & 78.37 & 78.68 & 76.09 \\
        Spotify & 98.85 & 95.61 & 96.24 & 91.71 & 98.68 & 95.04 & 96.43 & 92.04 & \underline{98.92} & 95.24 & 95.96 & 90.71 \\
        US Accidents Severity & \textbf{79.99} & 79.21 & 78.28 & \underline{79.60} & 79.43 & 79.53 & 79.45 & 79.08 & 79.58 & 78.83 & 77.75 & 78.47 \\
        Whisky & \underline{88.27} & 86.96 & 85.10 & 86.75 & 87.93 & 86.56 & 86.96 & 86.63 & 88.22 & 86.81 & 86.83 & 86.62 \\
        Yelp & 80.72 & 79.57 & 78.09 & 76.61 & 80.36 & 79.54 & 78.44 & 77.54 & \textbf{80.83} & 79.50 & 77.84 & 76.49 \\
        Zomato & \textbf{73.37} & 72.64 & 71.03 & 71.85 & 72.87 & 71.94 & 70.57 & 71.79 & \underline{73.10} & 71.53 & 70.88 & 71.46 \\
        \bottomrule
    \end{tabular}}
\end{table}

\begin{table}[h]
    \caption{\textbf{Classification results for CARTE datasets (part 3 of 3).} Raw ROC AUC scores multiplied by 100, averaged over 5 random seeds for a train size of 1,024 samples, for the remaining learners, fed with the cell-wise LLM encoder (Llama-3.1-8B) or TF-IDF, and the methods with internal string handling; TabLLM has no result on two datasets (--). Best in bold, second best underlined, both computed over all 36 methods.}
    \label{tab:carte_clf_3}
    \centering
    \scriptsize
    \setlength{\tabcolsep}{3.5pt}
    \resizebox{\linewidth}{!}{%
    \begin{tabular}{l|cc|ccc|cc|cc|c|c|c}
        \toprule
        \textbf{Method} & \multicolumn{2}{c}{\textbf{XGBoost (tuned)}} & \multicolumn{3}{c}{\textbf{CatBoost}} & \multicolumn{2}{c}{\textbf{RealMLP (tuned)}} & \multicolumn{2}{c}{\textbf{Ridge}} & \textbf{ContextTab} & \textbf{TabSTAR} & \textbf{TabLLM} \\
        \textbf{String encoding} & LLM encoder & TF-IDF & LLM encoder & TF-IDF & internal & LLM encoder & TF-IDF & LLM encoder & TF-IDF & internal & internal & internal \\
        \midrule
        Chocolate Bar Ratings & 77.68 & 78.97 & 77.69 & 78.55 & 70.29 & 77.11 & 78.27 & 77.89 & 77.53 & 75.65 & 80.92 & 82.30 \\
        Coffee Ratings & 90.30 & 91.88 & 90.42 & 92.34 & 74.94 & 90.90 & 92.25 & 89.29 & 92.38 & 82.71 & \underline{93.75} & \textbf{95.45} \\
        Michelin & 88.31 & 87.50 & 88.25 & 87.13 & 84.77 & 88.09 & 86.55 & 88.25 & 86.25 & 86.71 & 86.58 & 87.10 \\
        NBA Draft & 88.13 & 88.25 & 86.71 & 85.89 & 88.17 & 87.87 & 87.81 & 86.39 & 88.04 & 88.27 & 88.19 & 85.90 \\
        Ramen Ratings & 68.95 & 68.10 & 68.26 & 68.08 & 67.74 & 68.85 & 67.71 & 69.07 & 68.88 & 68.68 & \textbf{79.11} & 67.37 \\
        Roger Ebert & 78.09 & 75.42 & 76.76 & 74.58 & 75.31 & 76.79 & 74.05 & 76.22 & 75.77 & 75.49 & 76.21 & 76.28 \\
        Spotify & 94.22 & 88.92 & 94.39 & 89.85 & 90.13 & 93.53 & 89.06 & 94.52 & 86.66 & 93.93 & 91.72 & 96.27 \\
        US Accidents Severity & 76.94 & 76.90 & 75.56 & 76.82 & 78.76 & 75.47 & 77.26 & 76.64 & 75.97 & 79.49 & 74.11 & 74.46 \\
        Whisky & 81.41 & 85.52 & 84.07 & 85.29 & 84.95 & 75.74 & 86.18 & 86.42 & 86.53 & 86.47 & 85.10 & 84.10 \\
        Yelp & 75.73 & 74.48 & 74.87 & 73.14 & 73.33 & 73.90 & 72.56 & 75.51 & 72.45 & 76.51 & 73.97 & 73.72 \\
        Zomato & 69.39 & 70.47 & 69.64 & 70.25 & 67.16 & 66.34 & 70.02 & 71.07 & 71.71 & 71.27 & 67.27 & 68.57 \\
        \bottomrule
    \end{tabular}}
\end{table}

\FloatBarrier

\begin{table}[h]
    \caption{\textbf{Classification results for WikiDBs datasets (part 1 of 3).} Raw ROC AUC scores multiplied by 100, averaged over 5 random seeds for a train size of 1,024 samples, for TabPFN-2.5, TabICLv2 and TabPFN-3, each fed with TILE, CASE-like, the cell-wise LLM encoder (Llama-3.1-8B) or TF-IDF. Best in bold, second best underlined, both computed over all 36 methods.}
    \label{tab:wdbs_clf_1}
    \centering
    \scriptsize
    \setlength{\tabcolsep}{3.5pt}
    \resizebox{\linewidth}{!}{%
    \begin{tabular}{l|cccc|cccc|cccc}
        \toprule
        \textbf{Method} & \multicolumn{4}{c}{\textbf{TabPFN-2.5}} & \multicolumn{4}{c}{\textbf{TabICLv2}} & \multicolumn{4}{c}{\textbf{TabPFN-3}} \\
        \textbf{String encoding} & TILE & CASE-like & LLM encoder & TF-IDF & TILE & CASE-like & LLM encoder & TF-IDF & TILE & CASE-like & LLM encoder & TF-IDF \\
        \midrule
        Artists Copyright & 80.23 & 74.83 & 78.61 & 78.53 & 80.53 & 75.09 & 78.60 & 77.12 & 79.72 & 75.16 & 76.22 & 77.36 \\
        Artworks Catalog & 97.45 & 97.23 & 97.21 & 97.31 & 97.61 & 97.38 & 97.37 & 97.09 & 97.11 & 97.29 & 97.00 & 96.99 \\
        CC Authors & 99.75 & 99.44 & 99.31 & 91.77 & 99.74 & 99.28 & 99.33 & 92.86 & 99.73 & 99.43 & 99.14 & 92.29 \\
        Defenders & 99.40 & 97.80 & 99.01 & 95.23 & \underline{99.41} & 97.74 & 99.04 & 95.75 & 99.37 & 97.76 & 99.02 & 94.83 \\
        Forward Players & 89.26 & 87.13 & 88.96 & 88.80 & 89.46 & 87.57 & 89.08 & 88.55 & 89.07 & 87.10 & 88.25 & 88.06 \\
        Geographers & 99.54 & 98.38 & 99.12 & 93.62 & 99.56 & 98.49 & 99.12 & 93.95 & 99.54 & 98.46 & 99.07 & 92.08 \\
\bottomrule
    \end{tabular}}
\end{table}

\begin{table}[h]
    \caption{\textbf{Classification results for WikiDBs datasets (part 2 of 3).} Raw ROC AUC scores multiplied by 100, averaged over 5 random seeds for a train size of 1,024 samples, for TabPFN-3.5, EXAONE-Tabular and Causilo, each fed with TILE, CASE-like, the cell-wise LLM encoder (Llama-3.1-8B) or TF-IDF. Best in bold, second best underlined, both computed over all 36 methods.}
    \label{tab:wdbs_clf_2}
    \centering
    \scriptsize
    \setlength{\tabcolsep}{3.5pt}
    \resizebox{\linewidth}{!}{%
    \begin{tabular}{l|cccc|cccc|cccc}
        \toprule
        \textbf{Method} & \multicolumn{4}{c}{\textbf{TabPFN-3.5}} & \multicolumn{4}{c}{\textbf{EXAONE-Tabular}} & \multicolumn{4}{c}{\textbf{Causilo}} \\
        \textbf{String encoding} & TILE & CASE-like & LLM encoder & TF-IDF & TILE & CASE-like & LLM encoder & TF-IDF & TILE & CASE-like & LLM encoder & TF-IDF \\
        \midrule
        Artists Copyright & \underline{80.61} & 76.89 & 77.06 & 78.61 & 80.26 & 76.81 & 79.13 & 78.56 & \textbf{80.72} & 77.36 & 78.43 & 78.48 \\
        Artworks Catalog & 97.60 & \textbf{97.72} & 97.57 & 97.68 & 97.55 & 97.67 & 97.56 & 97.63 & \underline{97.70} & 97.66 & 97.48 & 97.45 \\
        CC Authors & \textbf{99.79} & 99.54 & 99.38 & 93.28 & 99.72 & 99.48 & 99.28 & 93.04 & \underline{99.75} & 99.46 & 99.31 & 93.02 \\
        Defenders & 99.41 & 98.56 & 99.12 & 96.06 & 99.36 & 98.23 & 99.13 & 95.91 & \textbf{99.47} & 98.14 & 99.08 & 95.55 \\
        Forward Players & 89.67 & 89.25 & 88.96 & 89.02 & \underline{89.72} & 88.53 & 89.46 & 89.07 & \textbf{89.76} & 88.35 & 89.26 & 88.95 \\
        Geographers & \underline{99.57} & 98.83 & 98.97 & 95.45 & 99.49 & 98.73 & 99.11 & 94.17 & \textbf{99.57} & 98.67 & 99.07 & 94.12 \\
\bottomrule
    \end{tabular}}
\end{table}

\begin{table}[h]
    \caption{\textbf{Classification results for WikiDBs datasets (part 3 of 3).} Raw ROC AUC scores multiplied by 100, averaged over 5 random seeds for a train size of 1,024 samples, for the remaining learners, fed with the cell-wise LLM encoder (Llama-3.1-8B) or TF-IDF, and the methods with internal string handling; TabLLM has no result on two datasets (--). Best in bold, second best underlined, both computed over all 36 methods.}
    \label{tab:wdbs_clf_3}
    \centering
    \scriptsize
    \setlength{\tabcolsep}{3.5pt}
    \resizebox{\linewidth}{!}{%
    \begin{tabular}{l|cc|ccc|cc|cc|c|c|c}
        \toprule
        \textbf{Method} & \multicolumn{2}{c}{\textbf{XGBoost (tuned)}} & \multicolumn{3}{c}{\textbf{CatBoost}} & \multicolumn{2}{c}{\textbf{RealMLP (tuned)}} & \multicolumn{2}{c}{\textbf{Ridge}} & \textbf{ContextTab} & \textbf{TabSTAR} & \textbf{TabLLM} \\
        \textbf{String encoding} & LLM encoder & TF-IDF & LLM encoder & TF-IDF & internal & LLM encoder & TF-IDF & LLM encoder & TF-IDF & internal & internal & internal \\
        \midrule
        Artists Copyright & 75.88 & 75.59 & 76.87 & 75.43 & 74.86 & 75.21 & 75.94 & 74.56 & 73.73 & 76.92 & 73.27 & 75.68 \\
        Artworks Catalog & 96.38 & 95.91 & 96.94 & 96.57 & 93.54 & 95.90 & 96.35 & 93.31 & 93.92 & 96.82 & 93.45 & 96.25 \\
        CC Authors & 98.48 & 87.91 & 97.96 & 87.95 & 83.69 & 98.96 & 85.61 & 98.99 & 80.54 & 97.86 & 98.85 & 99.29 \\
        Defenders & 98.66 & 93.47 & 98.91 & 94.11 & 88.96 & 98.64 & 93.62 & 98.59 & 90.19 & 97.30 & 96.61 & 98.58 \\
        Forward Players & 88.68 & 87.67 & 88.87 & 87.64 & 85.98 & 88.05 & 87.24 & 87.22 & 86.18 & 88.73 & 84.08 & 88.61 \\
        Geographers & 98.70 & 91.71 & 98.94 & 92.79 & 86.14 & 99.09 & 91.41 & 99.19 & 88.40 & 95.70 & 96.58 & 98.94 \\
\bottomrule
    \end{tabular}}
\end{table}

\FloatBarrier

\FloatBarrier

\clearpage
\section{Models, software and licenses}\label{app:licenses}

TILE is an architecture rather than a model, and thus relies on existing, publicly released artifacts. We list here everything we used, with its source, so that our results can be reproduced.

\paragraph{Language models} All backbones are downloaded from Hugging Face. From the Llama family \citep{grattafiori2024llama} we use Llama-3.1-70B and Llama-3.1-8B (\texttt{meta-llama/Llama-3.1-70B} and \texttt{meta-llama/Llama-3.1-8B}), under the Llama 3.1 Community License, and Llama-3.2-3B and Llama-3.2-1B (\texttt{meta-llama/Llama-3.2-3B} and \texttt{meta-llama/Llama-3.2-1B}), under the Llama 3.2 Community License. From the Qwen3 family \citep{yang2025qwen3} we use the three base checkpoints \texttt{Qwen/Qwen3-8B-Base}, \texttt{Qwen/Qwen3-4B-Base} and \texttt{Qwen/Qwen3-0.6B-Base}, released under the Apache 2.0 license. Llama-3.1-8B is the default backbone of TILE, and is also the model behind the cell-wise LLM encoder baseline.

\paragraph{Table foundation models and tabular learners} TILE and the encoding baselines are evaluated on top of TabPFN-2.5 \citep{grinsztajn2025tabpfn} and TabPFN-3 \citep{grinsztajn2026tabpfn}, from Prior Labs (\url{https://github.com/PriorLabs/TabPFN}), and TabICLv2 \citep{qu2026tabiclv2} (\url{https://github.com/soda-inria/tabicl}). The other learners are RealMLP \citep{holzmuller2024better}, used through \texttt{pytabkit} (\url{https://github.com/dholzmueller/pytabkit}), XGBoost \citep{chen2016xgboost}, CatBoost \citep{prokhorenkova2018catboost}, and Ridge regression from \texttt{scikit-learn}.

\paragraph{End-to-end baselines} ContextTab \citep{spinaci2025contexttab} is used through SAP's \texttt{sap-rpt-oss} release (\url{https://github.com/SAP-samples/contexttab}), TabSTAR \citep{arazi2025tabstar} through the \texttt{tabstar} package (\url{https://github.com/eyalbd2/TabSTAR}), and TabLLM \citep{hegselmann2023tabllm} through the authors' code together with the t-few recipe it builds on; we run it with T0-3B (\texttt{bigscience/T0\_3B}).

\paragraph{Software} The string-encoding baselines and all table preprocessing use \texttt{skrub} (\url{https://skrub-data.org}), whose \texttt{StringEncoder}, \texttt{LLMEncoder} and \texttt{TableVectorizer} we rely on throughout, together with \texttt{scikit-learn}, \texttt{sentence-transformers} for the cell-wise embeddings, and PyTorch and Hugging Face \texttt{transformers} for the LLM forward passes.

Apart from the Llama checkpoints, which carry the community licenses quoted above, every model and library listed here is distributed under a permissive open-source license (Apache 2.0, BSD-3-Clause or MIT). We use each of them for research purposes and in accordance with its terms.

\end{document}